\documentclass[sigconf]{acmart}

\usepackage{booktabs} 
\usepackage{amsmath}
\usepackage{wrapfig}
\usepackage{subfigure}
\graphicspath{ {figures/} }
\usepackage{xcolor}
\usepackage{tabularx}

\usepackage[linesnumbered,ruled,vlined]{algorithm2e}
\usepackage[noend]{algpseudocode}
\SetKwInput{KwInput}{Input}              
\SetKwInput{KwOutput}{Output}
\SetSideCommentLeft
\SetFillComment

\begin{document}
\title{AutoAI-TS: AutoAI for Time Series Forecasting}

\author{Syed Yousaf Shah, Dhaval Patel, Long Vu, Xuan-Hong Dang, Bei Chen, Peter Kirchner, Horst Samulowitz, David Wood, Gregory Bramble, Wesley M. Gifford, Giridhar Ganapavarapu, Roman Vaculin, and  Petros Zerfos}
\affiliation{
  \institution{IBM Thomas J. Watson Research Center}
   \city{Yorktown Heights}
   \state{NY}
  \postcode{10598}
}
\email{syshah@us.ibm.com, pateldha@us.ibm.com, lhvu@us.ibm.com, Xuan-Hong.Dang@ibm.com}
\email{beichen2@ie.ibm.com, pdk@us.ibm.com, samulowitz@us.ibm.com, dawood@us.ibm.com, gbramble@us.ibm.com}
\email{wmgifford@us.ibm.com, Giridhar.Ganapavarapu@ibm.com, vaculin@us.ibm.com, pzerfos@us.ibm.com}
%
\renewcommand{\shortauthors}{S. Yousaf Shah et al.}


\begin{abstract}

A large number of time series forecasting models including traditional statistical models, machine learning models and more recently deep learning have been proposed in the literature. However, choosing the right model along with good parameter values that performs well on a given data is still challenging. Automatically providing a good set of models to users for a given dataset saves both time and effort from using trial-and-error approaches with a wide variety of available models along with parameter optimization. We present \emph{AutoAI for Time Series Forecasting (AutoAI-TS)} that provides users with a  zero configuration (\emph{zero-conf}) system to efficiently train, optimize and choose best forecasting model among various classes of models for the given dataset. With its flexible zero-conf design, AutoAI-TS automatically performs all the data preparation, model creation, parameter optimization, training and model selection for users and provides a trained model that is ready to use. For given data, AutoAI-TS utilizes a wide variety of models including classical statistical models, Machine Learning (ML) models, statistical-ML hybrid models and deep learning models along with various transformations to create forecasting pipelines. It then evaluates and ranks pipelines using the proposed T-Daub mechanism to choose the best pipeline. The paper describe in detail all the technical aspects of AutoAI-TS along with extensive benchmarking on a variety of real world data sets for various use-cases. Benchmark results show that \emph{AutoAI-TS}, with no manual configuration from the user, automatically trains and selects pipelines that on average outperform existing state-of-the-art time series forecasting toolkits.

\end{abstract}



\keywords{machine learning; ML pipelines; AutoML; time series; optimization}

\maketitle

%
%


\section{Introduction}				\label{sec:intro}


The task of constructing and tuning machine learning pipelines \cite{kegl2017, ayria2018, zhou2018} for a given problem and data set is a complex, manual, and labor-intensive endeavor that requires a diverse team of data scientists and domain knowledge (subject matter) experts. It is an iterative process that involves design choices and fine tuning of various stages of data processing and modeling, including data cleansing, data imputation, feature extraction and engineering, model selection and parameter tuning. Automation of AI and Machine Learning \cite{elshawi2019, he2019}, usually referred to as AutoAI \& AutoML, aims to provide automated assistance for this complex process that requires highly specialized and hard-to-find data scientists and professionals.

AutoML has emerged as the natural evolution from earlier efforts on automated hyper-parameter optimization (HPO). It has gained increased popularity in recent years, including commercial offerings by the major cloud service providers \cite{googleAutoML, ibmAutoAI} and leading open source ML libraries \cite{h2oAutoML}. However, the overwhelming amount of work, including on benchmarking frameworks and state-of-the-art comparisons~\cite{benchmark2019automl}, has focused so far on the classification task. We believe that it's due time to recalibrate the attention of AutoAI/AutoML research efforts to the equally important domain of time series analysis, starting with the ever popular task of time series forecasting.


Automating the model selection and tuning of time series forecasting algorithms has been extensively studied in the context of model selection {\em within} individual classes of models such as auto-regressive/ARIMA~\cite{pmdarima} with the Box-Jenkins methodology, exponential smoothing models with the state-space methodology~\cite{Hyndman2}, as well a newer efforts such as DeepAR~\cite{deepAR} that augment deep neural networks and auto-regressive capabilities. However, due to the variety and heterogeneity of stochastic processes that generate time series data, stemming from diverse domains such as Internet-of-Things sensor measurements, financial data series, cloud application and service monitoring data to name a few, as well as the proliferation of time series forecasting methods~\cite{makridakis2018}, ranging from the classical statistical time series models, to ML-based techniques, to deep neural network-based ones~\cite{lai2017}, there is an ever expanding need to automate pipeline selection and tuning {\em across} model classes and approaches. 
The work in this paper aims to cover such need through a zero configuration (\emph{zero-conf})  system framework that encompasses many of the aforementioned approaches. Interestingly, and as we further validated during the experimentation efforts of our system, no single time series forecasting technique from those tested is able to accurately capture the complex patterns that are present in time series data from all domains that were analyzed.

We present the design, implementation and performance evaluation of a system called \emph{AutoAI for Time Series Forecasting (AutoAI-TS)} for the automation of time series forecasting process using a variety of models. The goal of AutoAI-TS is to be a \emph{zero-conf} system, wherein the user simply drops-in their data set and the system transparently performs all the complex tasks of feature engineering, training, parameter tuning, model ranking and returns one or more of the best performing trained models ready for prediction. A single set of {\em scikit-learn}~\cite{sklearn_api} compatible APIs provides a unified view of the AutoAI-TS system to the end users contributing to ease of use and familiarity of interfaces. Our goal is to design and develop a system that is able to achieve high accuracy on the time series forecasting task with minimal to no manual user input. 

Our architecture incorporates various classes of time series forecasting models incorporated from open-source domain as well as set of models developed in-house with {\em scikit-learn} compatible APIs. The system also utilizes data transformations along with forecasting model to create pipelines and evaluate their feasibility for the given data. We further tailored hyper-parameter optimization techniques for each class of models included in our system and introduced a new technique for automatic (dynamic) extraction of optimal look-back window length for ML models, based on input data. The list below summarizes our main contributions in this paper
\begin{enumerate}
\item Design and implementation of a \emph{zero-conf} and \textbf{\textit{extensible AutoAI-TS system framework}}, which develops and integrates, as one uniform system, several classes of time series forecasting models ranging from Holt-Winters Seasonal Additive and Multiplicative, ARIMA, BATS~\cite{BATS} (by Livera, Hyndman and Snyder), Random Forest Regression, Support Vector Regression, XGBoost, Linear Regression, etc.
\item Designed and implemented an automatic pipeline ranking and selection technique, namely \textbf{\textit{T-Daub with \emph{reverse} progressive data allocation}}, which makes it suitable for pipeline evaluation on time series data that exhibits sequential dependencies.
\item Devised a novel \textbf{\textit{automatic look-back window length estimation}} mechanism for the ML-based pipelines.
\item Extensive \textbf{\textit{performance evaluation results on \texttt{62+} univariate and \texttt{9+} multivariate data sets}} sourced from a variety of application domains and ranging in size from a few dozen samples to more than \texttt{1,400,000}. We demonstrate performance of our AutoAI-TS system without any manual tuning and with reasonable training time. We further provide comparison results to \texttt{10} more state-of-the-art approaches for automated time series forecasting from the research literature and open source domain. 
\end{enumerate}
The remainder of this paper is organized as follows: Section \ref{sec:background} provides brief background and discussion of related work. Section \ref{sec:autoaitsdata} discusses the implications of time series data when automation of pipelines is introduced, whereas Section \ref{sec:sysarch} provides details on the AutoAI-TS system architecture. Section \ref{sec:results} elaborates on our extensive experiments and the performance evaluation, with details on datasets and results, including comparison to other state-of-the-art systems. Section \ref{sec:conclusion} concludes this paper.

\section{Background and Related Work}
\label{sec:background}

\emph{AutoAI TS} provides an end-to-end solution starting from the input time series to the deployment of the top performing prediction pipelines. The development of fully automated time series forecasting tools has been increasingly attracting attention in recent years due to their wide application across industries, e.g., financial services, retail, and manufacturing. There are a number open source software tools proposed in literature (\cite{HyndmanKhandakar:2008}, \cite{sktime}, \cite{Taylor2017}, \cite{h2oAutoML}
) and commercial products (e.g, \texttt{DataRobot}, \texttt{Oneclick}, \texttt{Azure Time Series Insights}, \texttt{Element AI}). Some of the early works rely on the selection of a single statistical model. For example, Hyndman and his team have developed a set of algorithms for fitting ARIMA and extended exponential smoothing methods for automatic selection of the model orders and parameters \cite{HyndmanKhandakar:2008, Hyndman2}. These methods are available via open source R package \texttt{forecast} and Python library \texttt{Pyramid}. \texttt{Prophet} from Facebook automates the time series forecasting based on the Generalized Additive Model \cite{Hast:Tibs:1987}. The fundamental algorithm assumes the underlying data can be systematically decomposed into trend, seasonality and causal effects \cite{Taylor2017}. \texttt{Prophet} targets big data applications and is available both in R and Python. In recent literature the efforts are more focused on automating machine learning (ML) and deep learning (DL) algorithms for time series prediction. One of the challenges of Auto-ML approaches lies in the hyper-parameter tuning. Google \texttt{Vizier} \cite{GoogleVizier} provides a automatic black-box optimization for ML model tuning. \cite{sktime} offers a unified interface for ML pipelines for time series prediction which compatible with scikit-learn. \texttt{DeepAR} by Amazon proposes to apply deep autoregressive recurrent networks for probability forecasting \cite{deepAR} of time series data. \texttt{N-BEATS} by Element AI \cite{Oreshkin2020:N-BEATS} proposes a hybrid model that uses deep learning and autoregressive models for building an accurate and explainable time series forecasting model. In comparison to the aforementioned tools, instead of relying on one type of model, \emph{AutoAI-TS} provides a wide range of statistical, ML, statistical-ML hybrid and DL pipelines. Joint optimization enables \emph{AutoAI-TS} to quickly identify the top performing pipelines at scale. 

\section{Automation of AI for Time Series Data}
\label{sec:autoaitsdata}
In this paper, we focus on automation of AI for time series data and we aim to automate available time series forecasting techniques and models with minimum possible input from user. The overall \emph{AutoAI-TS} architecture is aimed to provide unified APIs and user experience across multiple classes of forecasting models as well as architectures. We utilize \emph{scikit-learn} style APIs, as they are widely used for allowing flexibility and compatibility with similar systems. However, since time series data differs from independent and identically distributed (IID) data in many ways; characteristics of time series data such as causality, sampling frequency, trend, seasonality, non-stationarity, prediction horizon, etc., need to be modeled in the time series forecasting process. Therefore, we need a forecasting system that can address time series specific requirements. 


%
%



\emph{AutoAI-TS} encompasses the family of classical statistical forecasting models including ARIMA, ARMA, Additive and Multiplicative Triple Exponential Smoothing also known as Holt-winters and BATS (Box-Cox transformation, ARMA errors, Trend and Seasonal components) model  \cite{BATS}, that we implemented for efficient, parallel and automatic search of corresponding model parameters. We also include Machine Learning models (ML) such as Random-Forest, XGBoost, Linear Regression, SGD Regression, and  Statistical-ML Hybrid Models and Deep Learning based models, which we either implemented or incorporated from existing open source implementations. Generally, ML based approaches perform transformations on time series data and then model time series forecasting problem as an IID problem, such as classification. More specifically, input time series data is first transformed using stateless transformer (transformers that do not remember the state of the operation) such as \texttt{log}, \texttt{fisher}, \texttt{box\_cox}, etc. Then, stateful transformations are optionally performed, stateful transformations retain the knowledge of the sequence of operation that are performed such as Difference, Flatten, Localized Flatten and Normalized Flatten. Finally, the transformed data is passed to ML models for training. At prediction time, we need to reverse transform the data output from the model to the original form and scale. Therefore, inverse transformations are applied in the reverse order of application, i.e., the stateful inverse transformation followed by stateless inverse transformation \cite{DBLP:conf/icdcs/IyengarKPRS19} \cite{DBLP:conf/icws2/ShrivastavaPGSK19}.


All these model families for prediction operate differently, have different requirements and sets of parameters that need to be optimized in order for these models to produce good forecasts. Since, there is no single model that can perform well across diverse real-world datasets, having heterogeneous models enables capturing the inherent characteristics of these data sets. Our motivation for having a diverse set of pipelines is to be able to model various data sets, as well as address various user requirements. 
Different users have different requirements for their data analytics task, e.g., compute time, size of data to be analyzed, how far in future to predict, etc. Some users might prefer and use batch processing hence allowing more time to train and optimize multiple models to get better accuracy; others might be constrained by time and compute resources preferring models that can be trained quickly with relatively low compute power. Such distinction is very useful in cases where users train models and produce forecasts on their local machines, as they can choose models that suite their requirements. However, supporting diverse set of pipelines makes it challenging to provide a unified view to the user such that all the complexities and models heterogeneity is transparent to the end user and all models operate in same way from user perspective.

\begin{figure}[t]
\centering
\includegraphics[height=1.6in]{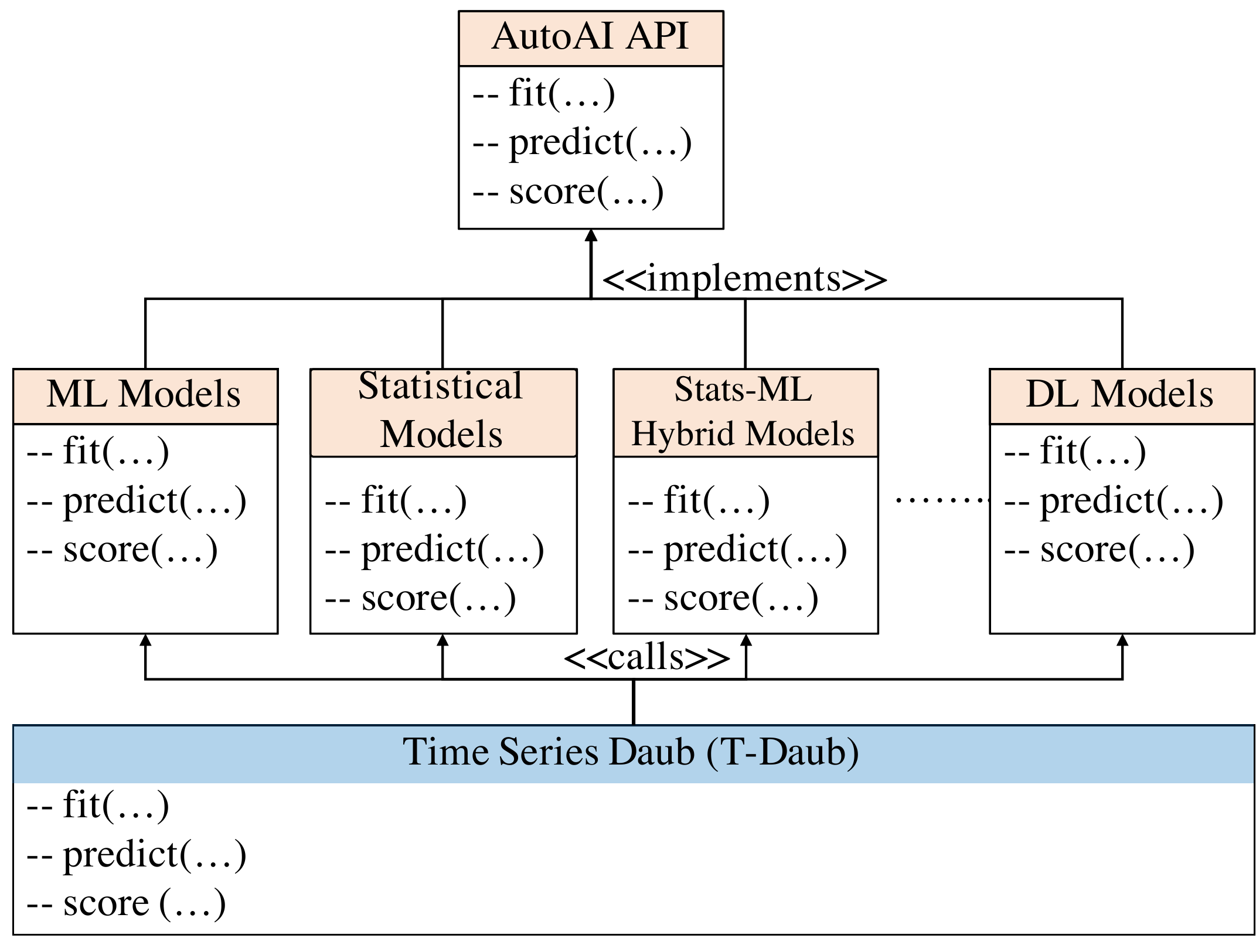}
\caption{AutoAI Time Series API Implementation Structure}
\label{fig:api}
\end{figure}

As first-step towards enabling transparent access to heterogeneous models is to have common APIs, we follow \emph{scikit-learn} approach to define APIs for our system. This ensures that our models and transformation are easily pluggable making it easier to extend the framework with new models and capabilities. Figure \ref{fig:api} shows the API layout of our framework, the assets in our framework implement estimators or transformers APIs and are joined together in a \emph{scikit-learn } style pipeline. Each model implements \texttt{fit}, \texttt{predict} and optionally \texttt{score} methods whereas transformers implement \texttt{transform} methods. T-Daub which is our pipeline selection mechanism is also implemented as an estimator which invokes  \texttt{fit}, \texttt{predict} methods of the corresponding pipelines. Pipelines encapsulate all the complexities and perform all necessary tasks internally, such as model parameter search and data reshaping, etc. All transformers and estimators are designed to use same input output format and shape to ensure compatibility. We use \textit{2D array} as input and output schema for pipelines, estimators and transformers. For example, \texttt{fit} and \texttt{predict} expect a \textit{2D array} in which columns represent different time series and rows represent samples. The \texttt{predict} function produces output in form of a \textit{2D array} in which columns correspond to input time series and rows are number of future values predicted (prediction horizon). Each component can internally manipulate data but for simplified user experience, we use a \textit{2D array} which user can easily prepare and provide to the pipeline. The pipelines are ranked and selected based on their predicted performance using the T-Daub algorithm explained in section \ref{sec:tdaub}.



%
%
%
%
%
%
%
%
%

\section{System Architecture}
\label{sec:sysarch}

\begin{figure}[t]
\centering
\includegraphics[width=2.9in]{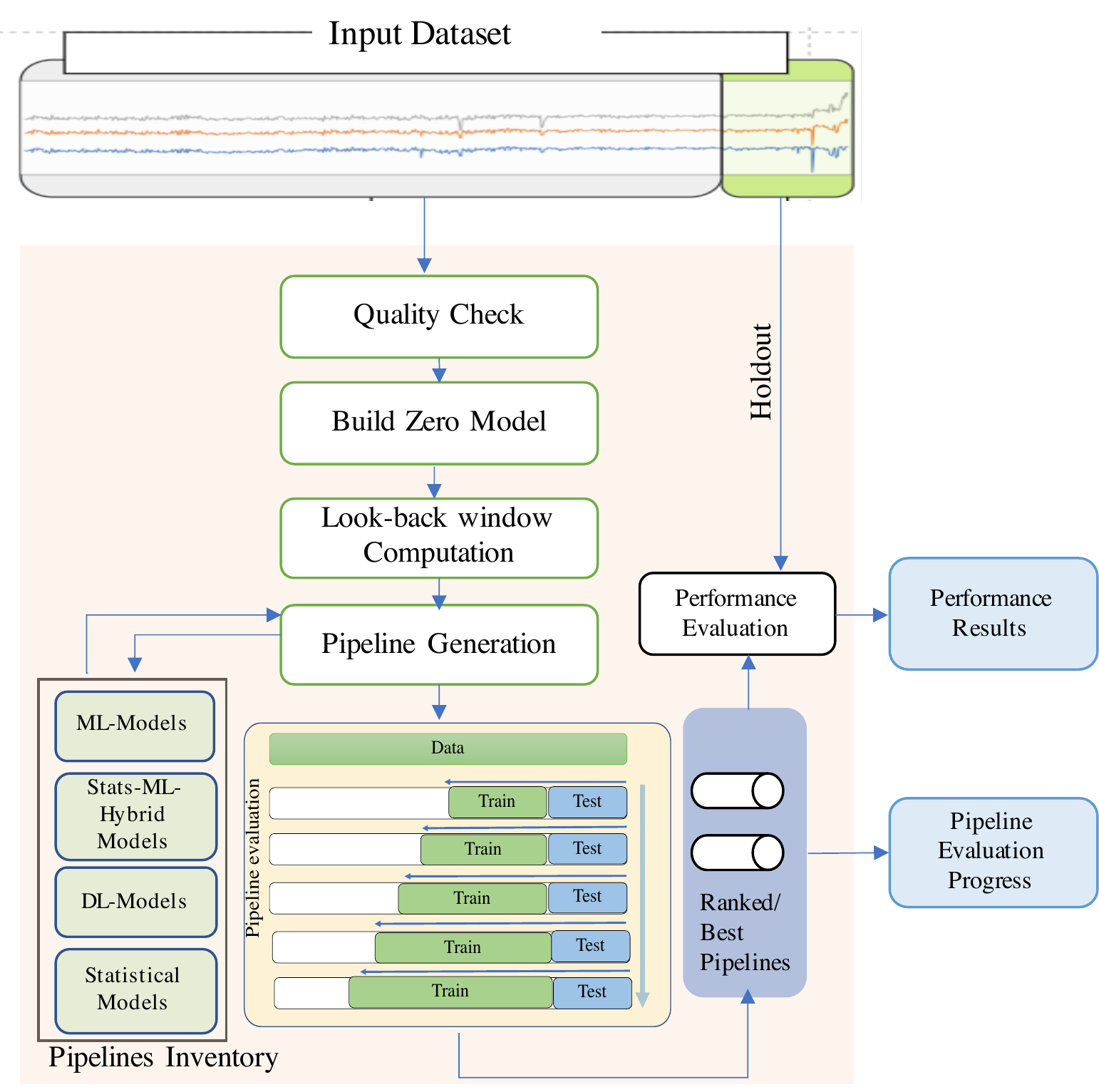}
\caption{\emph{AutoAI-TS} Overall Architecture}
\label{fig:autoaits_arch}
\vspace*{-0.20cm}
\end{figure}

Section \ref{sec:autoaitsdata} describes in detail the APIs and data semantics of inputs and outputs for functions. This section describes the overall system architecture and how different components of the system are integrated together and function to run and train models for predictions on a given data set. Figure~\ref{fig:autoaits_arch} shows overall architecture of the system.

Once the data is provided to the system, it performs an initial quality check of the input data which includes looking for missing or \texttt{NaN} values, unexpected characters or values such as strings in the time series, it also checks if there are negative values so that system can disable certain transformations such as \texttt{log} transform, etc. Once the data passes the checks and basic data cleaning is performed the system trains a basic model; the \textit{Zero Model} shown in figure \ref{fig:autoaits_arch}. This \textit{Zero Model} almost immediately provides us with a baseline model that is available for use. The \textit{Zero Model} simply outputs the most recent value of a time series as the next prediction. For prediction horizons greater than \textit{1} the  most recent value is repeated.

Most ML models and deep learning models use a history or look-back window during training and inference. The look-back window length can have significant impact on model performance and needs to be changed based on input data. \emph{AutoAI-TS} has ML models and DL models that use look-back window as well, so it is important to automatically and efficiently compute the length of look-back window based on the input data. We propose a novel mechanism that analyzes the input data and generates look-back window sizes. Section \ref{sec:lookback} describes in detail the look-back discovery mechanism. If the user specifies look-back window size then the \textit{look-back window generation} is skipped, otherwise one of the automatically discovered look-back window sizes is used by \textit{ML Models} and \textit{Stats-ML Hybrid} models.

After look-back size has been decided, pipelines are generated from existing models and transformation. Currently, pre-composed pipelines are instantiated but the system can also dynamically generate new pipelines. In addition to better modeling of the data, these transformations can be used to fill-in missing values in data i.e., interpolator transformer can be used or up/down sampling transformers can be used for irregular samples. The system has various classes of models and some models require regular data (e.g., ARIMA) whereas other (e.g., BATS) do not, for models that require regular data, we can use up/down sampling as transformation in pipeline before feeding data to models that require regular data, thus system can readily handle such cases. Our architecture also enables to implement transformers called \textit{Detectors} that can capture various characteristics of data such as, presence of negative or missing values, irregularly spaced data etc., so appropriate transformations can be applied. Our architecture works on pipelines level rather than individual model level which makes it easier to handle such cases. Statistical models in our system automatically estimate coefficients and optimize parameters based on the input training data whereas other models use default model parameters and their internal parameter optimization which are mostly based on open source implementations are disabled.

These pipelines are provided to pipeline selection mechanism shown as \textit{Time Series Daub} in figure \ref{fig:autoaits_arch}. The \textit{Time Series Daub (T-Daub)} is given the training part of the input data and creates various splits from it and trains the pipelines on these data splits to approximate the accuracy of these pipelines on full input data. It then ranks these pipelines according to the approximate expected performance and one or more top performing pipelines are chosen, more details on T-Daub are provided in section \ref{sec:tdaub}. During T-Daub evaluation of pipelines, user is provided with the overall progress and performance of the evaluated pipelines, such progress is displayed on command line as well as on the web-UI. The performance numbers such as accuracy, training and test times for the best pipeline(s) on holdout data are reported to the user as well and as final step the best pipelines(s) are trained on entire training dataset so that the pipelines utilize all the input data for training and are ready for producing predictions.

The unified APIs and data semantics enable the system to handle various use cases without any modification. For example, the user can provide single or multiple time series as input and the system will generate outputs for all the input time series. Moreover, the user can also set  \texttt{prediction~horizon >= 1}, meaning the user can ask for future predictions \emph{1-step} ahead or multiple steps ahead for one or more time series. This enables users to run the system uniformly for various settings and get results according to their use-cases. The system is designed to incorporate any other type of model family without requiring any changes to the system as long as the new models implement the common APIs as explained in section \ref{sec:autoaitsdata}. We have tested the system with about \texttt{80} different pipelines including deep learning models and the system successfully selected the best pipeline independent of type and nature of underlying models.


\subsection{Look-back Window Computation}
\label{sec:lookback}

Machine learning models and deep learning models, such as those incorporated in \emph{AutoAI-TS}, require a history window in order to forecast future values. We refer to this history window as look-back window and its length has significant impact on performance of the models. Efficient value of look-back window length is somewhat dependent on the input data, therefore it can either be set based on prior knowledge about the data or should be discovered based on characteristics of the input data. \emph{AutoAI-TS} does not assume prior knowledge about input data, hence we propose and implement an automatic look-back window length discovery mechanism, which for given input data computes most suitable look-back window to be used by an deep learning and ML models for time series forecasting. The look-back window can be found using the timestamp column of the time series or the value column of the time series or combination of both timestamp and values. Most of the existing look-back discovery mechanisms either use timestamp or values to find look-back window length whereas our proposed mechanism use both and automatically decides how much of weight should be given to either one. 


Given a univariate data set, our mechanism first conducts timestamp index based assessment. This assessment identifies the temporal frequency of the observations using timestamp column e.g., observations on daily basis (\texttt{1D}) or weekly basis (\texttt{1W}), etc. Next, the mechanism discovers the seasonal periods using the frequency of the input data. In our case, seasonal period denotes the number of observation in each season and we intend to discover multiple seasonal periods. For example, if discovered data frequency is \texttt{1D}, the possible seasonal periods are \texttt{7} (\texttt{1W}), \texttt{30} (\texttt{1M}), \texttt{365.25} (\texttt{1Y}), and so on. In Table \ref{tab:map_data_sea_per}, we provide a sample of mapping from data frequency to a set of possible seasonal periods. In summary, the time index assessment generates a range of possible look-back windows based on just the timestamp column.

Next, we conduct value index assessment that primarily uses the concept of zero crossings and spectral analysis to infer look-back windows. We describe the process of extracting the look-back window using the concept of zero crossing frequency as follows: we obtain the mean adjusted time series (i.e., difference of \texttt{x} and the mean value of \texttt{x}, for a signal \texttt{x}) and find the indices where zero crossings happen (positive to negative and vice-versa). Then, the look-back is computed as the average distance between adjacent crossing points. In addition, we discover one look-back window for each of the seasonal periods discovered based on time index inspection using spectral analysis. Given a seasonal period, the spectral analysis method infers power for various frequency values. We select the frequency with the highest power, provided the frequency value is nonzero. In the latter case, we use the frequency associated with second largest power. The inverse value of the selected frequency is returned as a possible value of look-back w.r.t.\ the seasonal period.

\setlength{\tabcolsep}{3pt}
\begin{table}[t]
  \centering
  \caption{Mapping of Frequency of Data to Seasonal Period}
    \begin{tabular}{lllllll}
    \hline
    Mapping & Minutes & Hours & Day & Week & Month & Year \\ \hline
    Year & & & & & & 1 \\    
    Month & & & & & 1 & 12 \\    
    Week & & & & 1 & 4 & 52 \\    
    Day & & & 1 & 7 & 30 & 365.25 \\    
    Hours & & 1 & 24 & 168 & 720 & 8766 \\    
    Minutes & 1 & 60 & 1440 & 10080 & 43200 & 525960 \\
    Seconds & 60 & 3600 & 86400 & 604800 & 2592000 & 31557600 \\    
    \hline
    \hline
    \end{tabular}%
  \label{tab:map_data_sea_per}%
\end{table}%

Finally, we post-process all the discovered look-back windows in order to recommend the final set of look-back values. The process includes performing the following sanity checks:
\begin{itemize}
\item All look-back values greater than the length of the dataset are discarded.
\item Values greater than \texttt{max\_look\_back} (user defined input, if specified) are discarded along with values of \texttt{0} and \texttt{1}.
\item If no value is found then the default values passed to the function is returned (we use \texttt{8} as default value).
\item In case of multiple look-back windows, we order the values based on a preference criteria. In this paper, we compute an influence vector for each look-back window, where each index in influence vector is a performance measure computed from applying a  simple models on a subset of data, e.g. F-test from linear regression, mutual information based measure to capture any relationship, or mean absolute error of random forest model. Given a signal \texttt{x} and a look-back window \texttt{lw}, we randomly sample nearly 800 windows and obtain a dataset of X (\texttt{800 x lw}), y (\texttt{800 x 1}). Note that, y[i] is the next immediate observation of its corresponding window X[i]. Now, we calculate the performance measure using X and y for a given choice (i.e., F-test(X,y)). The influence vector is converted into an influence rank vector using respective index, and the average value of influence rank is used to sort the look-back index.
\end{itemize}

In the case of multivariate time series, we generate a set of recommended values for each time series separately by calling a univariate look-back discovery process. Next, we obtain the union of look-back values by selecting the one preferred value for each time series. Let, \texttt{lwset} denote the set of recommended values, containing one value (preferred value) from each time series. Thus, the maximum length of \texttt{lwset} is equal to the number of time series in the multivariate data set. The set \texttt{lwset} can be processed in two different ways. As option 1, we order look-back window values in \texttt{lwset} in decreasing order. The ordered look-back set is processed in the descending order. While processing a look-back window, \texttt{lw}, if the look-back value (\texttt{lw}) satisfies the following condition:
$$\texttt{lw} * \texttt{num\_timeseries} > \texttt{max\_look\_back}$$

We cap that value by \texttt{max}(\texttt{1},$\frac{\texttt{max\_look\_back}}{\texttt{num\_timeseries}}$) and add the new value to \texttt{selectedLW}. When the above condition is not satisfied, we add the original value (\texttt{lw}) to \texttt{selectedLW}. Once we process all the values we return the newly generated set. As an alternate option, we ignore all the values that violate above condition.

\subsection{Time Series Daub}
\label{sec:tdaub}

\begin{figure}[H]
\centering
\includegraphics[width=2.4in]{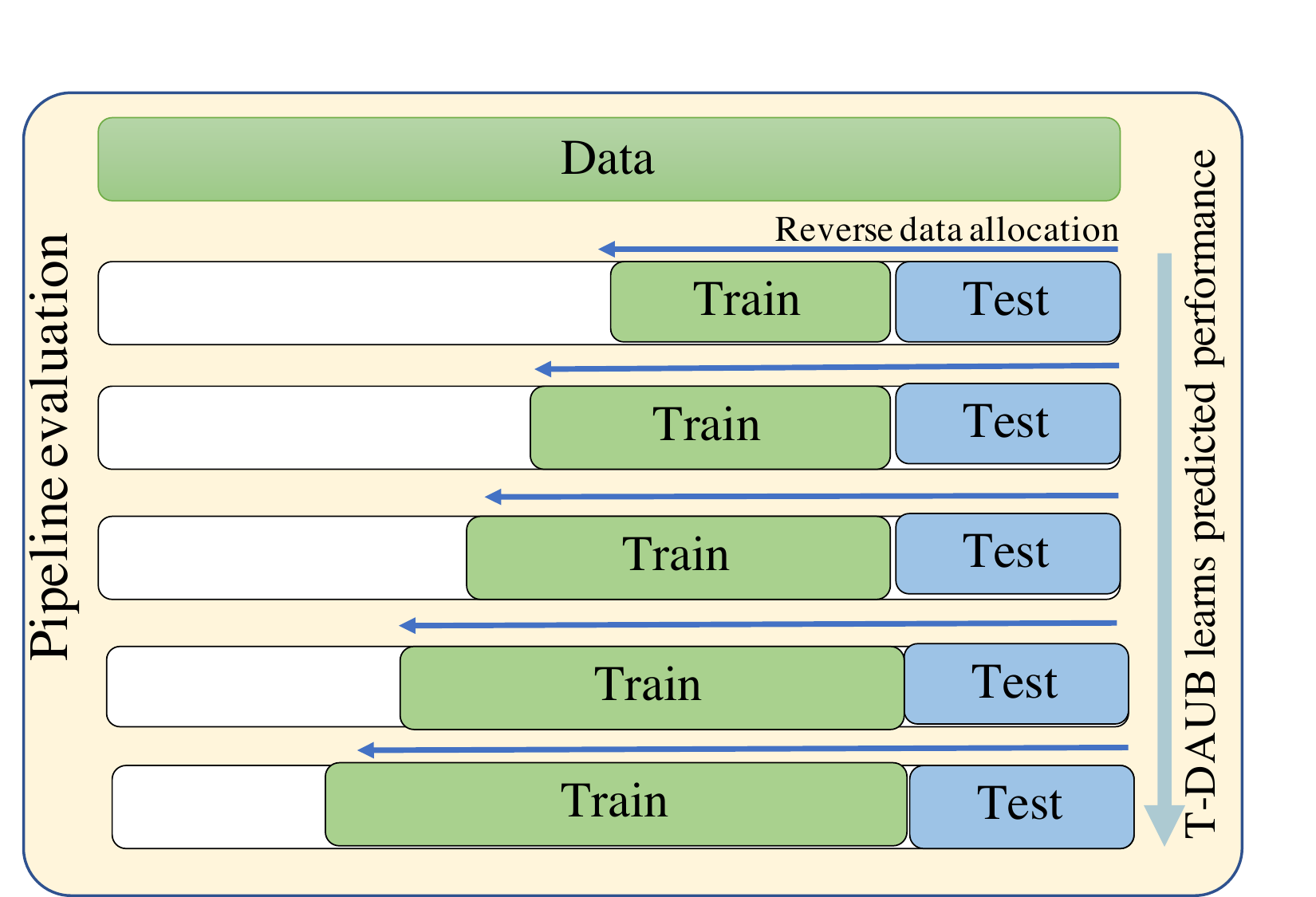}
\caption{T-Daub pipeline evaluation with most recent data splits (reverse allocation) for training}
\label{fig:tdaub}
\end{figure}

In \emph{AutoAI-TS}, we have a mix of pipelines and we need to find the best pipeline for the given data without running every pipeline on the full dataset. We use Time series Data Allocation Using Upper Bounds (T-Daub) as a pipeline ranking and selection mechanism which helps us discard unpromising pipelines in earlier stages of pipeline evaluation. T-Daub is inspired by Data Allocation Using Upper Bounds (Daub) \cite{Sabharwal2015SelectingNL} and is a modified version of the original \emph{Daub} to support time series data. T-Daub is a joint optimizer that sequentially allocates small subsets (called allocations) of training data, with latest data first, amongst a large set of forecasting pipelines. T-Daub itself is implemented as an estimator implementing APIs shown in figure \ref{fig:api}, it executes each of the pre-built pipelines by calling their \texttt{fit} method of these pipelines to train them. Pipeline execution and evaluation is based on a priority queue which ensures that the more promising pipeline is executed first. As opposed to the original Daub algorithm the data in T-Daub is allocated in reverse order i.e., starting from the most \emph{recent data} (see figure \ref{fig:tdaub}) to each pipeline. It is important to notice that in reverse allocation, all splits are created in reverse order which means that each allocation is created starting from the end of the training set and always contains the most recent data.

\noindent\textbf{Definitions of algorithm parameters}:  
\begin{itemize}
\item \texttt{min\_allocation\_size}: the smallest data chunk provided to pipelines, optionally set by user.
\item \texttt{allocation\_size}: the increment to the data to allocate size until the \texttt{fixed\_allocation\_cutoff} limit is reached.
\item \texttt{fixed\_allocation\_cutoff}: the limit up to which fixed size data chunks are allocated. Default is set to 5 times allocation\textunderscore size. 
\item \texttt{geo\_increment\_size}: geometric increment used as a multiplier after the fixed\_allocation\_cutoff limit is reached.
\item \texttt{run\_to\_completion}: the number of pipelines to run with all of the data before stopping. 
\end{itemize}
T-Daub executes only if length of dataset \texttt{T > min\_allocation\_size}. Otherwise, the data is given to all the pipelines and ranking is computed based on full data.

{\bf Algorithm 1} describes the T-Daub procedure. Given the training split of the original input time series data set \texttt{D} from the user, \texttt{D} is split into \texttt{\{T, H\}} in temporal order, where \texttt{T} is the training set and \texttt{H} is the holdout set. The data provided to T-Daub is the \texttt{80\%} training split of the overall data and does not include the holdout data that is used for final testing. The \texttt{20\%} holdout is only used for testing and reported evaluations, hence the \texttt{\{T, H\}} splits are created within \texttt{80\%} of the data. The algorithm consists of three major steps: 1) fixed allocation, where T-Daub is run on fixed size data allocations as shown by line-2 to line-8 in algorithm 1. 2) allocation acceleration in which data allocation size is increased geometrically (line-9 to line-17) and 3) scoring. The main idea is to rank the pipelines based on the projected learning curve in the fixed allocation part. Then the top ranked pipelines will acquire additional data in the allocation acceleration part for further training. Based on the scores achieved by pipelines in second step, top performing pipeline is selected and is trained on full training data. 
\\

\begin{algorithm}
\label{tdaub}
\caption{T-Daub}
\KwInput{pipelines $p_1...p_{np}$, \texttt{min\_allocation\_size}, \texttt{allocation\_size}, \texttt{fixed\_allocation\_cutoff}, \texttt{geo\_increment\_size}, \texttt{run\_to\_completion}}
\KwOutput{ranked pipelines}
\KwData{training set \texttt{T}}
\texttt{T} is split into \{\texttt{T$_1$}, \texttt{T$_2$}\}, \texttt{T$_1$} for training and \texttt{T$_2$} for testing.
\texttt{L}=length(\texttt{T$_1$})

\texttt{num\_fix\_runs} = int (\texttt{fixed\_allocation\_cutoff} / \texttt{min\_allocation\_size} )
%

 \tcp{Fixed allocation part}
\For{i in 1:\texttt{num\_fix\_runs}} 
{ \For{j in 1:np}
  { train $p_j$ on \texttt{T$_1$}[\texttt{L - min\_allocation\_size} $\times$ i +1 : \texttt{L}], test on \texttt{T$_2$} and compute score $s_{i,j}$. 
  }
}

\For{j in 1:np}
{ fit linear regression on \{$s_{1,j}$, \ldots, $s_{5,j}$\}, predict the score $s_j$ when data length = \texttt{L}. 
}

rank \{$s_{1}$, \ldots, $s_{np}$\} and the corresponding pipelines to obtain the ranked scores \{$s_{1}'$, \ldots, $s_{np}'$\} and the ranked pipeline \{$p_{1}'$, \ldots, $p_{np}'$\}

\tcp{Allocation acceleration part}
\tcc{In this part only the top pipelines get additional data allocation, which increases geometrically.}

\texttt{l} = \texttt{L - min\_allocation\_size} $\times$ num\textunderscore fix \textunderscore run

\texttt{next\_allocation} = int(\texttt{last\_allocation\_size} $\times$ \texttt{geo\_increment\_size} $\times$ \texttt{allocation\_size}$^{-1}$) $\times$ \texttt{allocation \_size}

\texttt{l} = \texttt{l} + \texttt{next\_allocation}

\While{\texttt{l < L}}
{ train the top pipeline $p_{1}'$ on\texttt{T$_1$}[\texttt{l}+1 : \texttt{L}], test on \texttt{T$_2$}, compute score and include it into the score vector of $p_{1}'$. \\
  Re-fit linear regression to predict $s_{1}'$ when data length = \texttt{L} \\
  Re-rank \{$s_{1}'$, \ldots, $s_{np}'$\} and \{$p_{1}'$, \ldots, $p_{np}'$\}\\
  \texttt{next\_allocation} = int(\texttt{last\_allocation\_size} $\times$ \texttt{geo\_increment\_size} $\times$ \texttt{allocation\_size}$^{-1}$) $\times$ \texttt{allocation\_size} \\
  \texttt{l} = \texttt{l} + \texttt{next\_allocation}
}
\tcp{T-Daub scoring} 

Train the top \texttt{run\_to\_completion} pipelines on \texttt{T$_1$} and compute the scores on \texttt{T$_2$}

Rank the score to obtain the final ranked list of pipelines. 
\end{algorithm}

\section{Benchmarking and Performance Evaluation}
\label{sec:results}

\begin{figure}[t]
\centering
\includegraphics[height=1in]{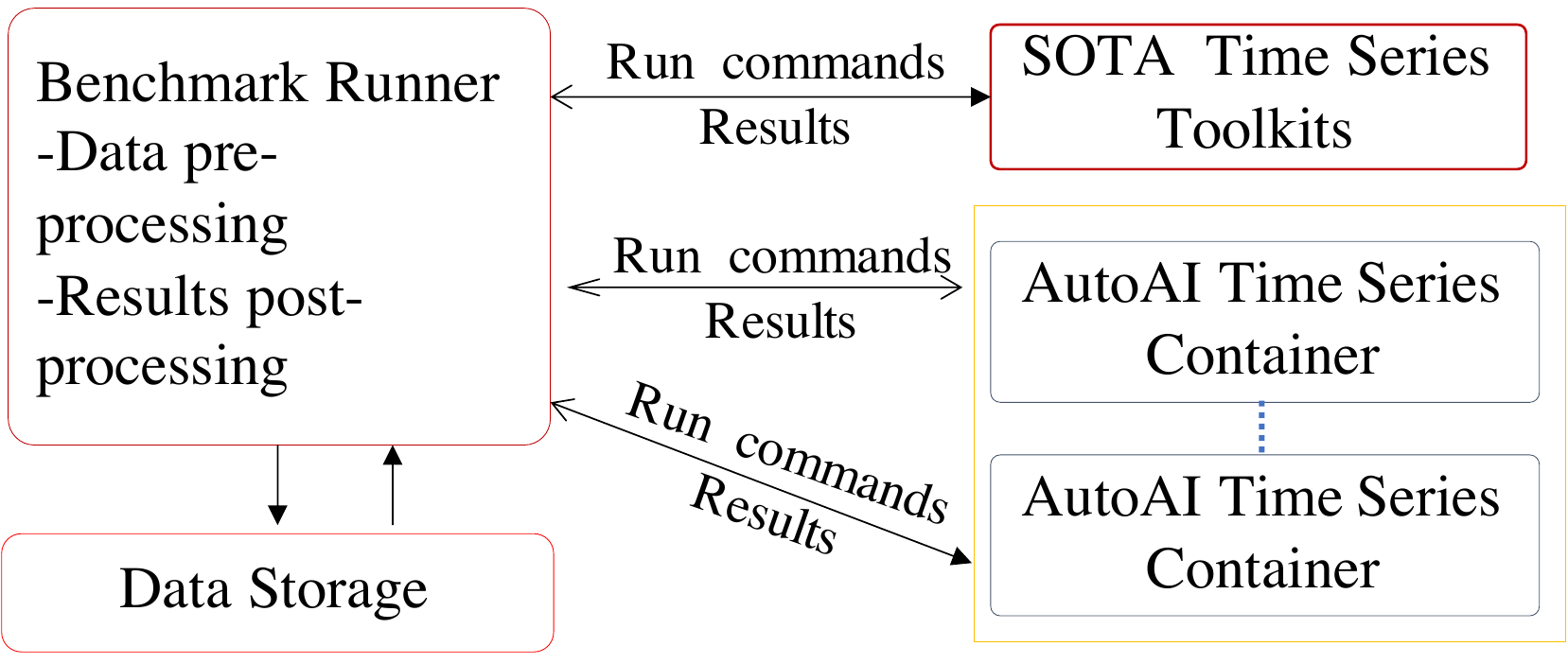}
\caption{AutoAI Time Series Benchmarking Framework}
\label{fig:bench}
\end{figure}

We have designed and developed a scalable container based benchmarking framework shown in figure \ref{fig:bench} in order to thoroughly and continuously test our overall system as well as individual pipelines. The benchmarking system automatically pulls the latest code from the repository and builds a container to run the benchmarking. The container reads data from a mapped disk cleans data and executes experiments. The benchmarking system also processes the output of the experiments and generates system performance reports. The benchmarking mechanism system also implements or imports existing implementations of the state-of-the-art (SOTA) time series toolkits which enables us to run experiments both on our system, i.e., \emph{AutoAI-TS} as well as on the \texttt{10} \emph{SOTA} frameworks with the same train-test split to get comparative performance results. We experimented with \emph{SOTA} frameworks such as, DeepAR (Amazon) \cite{salinas2020deepar}, Prophet (Facebook) \cite{taylor2018forecasting}, pmdarima (R-package in Python) \cite{pmdarima}, PyAF (python auto forecasting Open Source) \cite{PyAF}, Nbeats (Deep NN, 2020) \cite{Oreshkin2020:N-BEATS}, WindowRegressor  \cite{autoTS}, GLS  \cite{autoTS}, RollingRegressor  \cite{autoTS}, Motif  \cite{autoTS}, and Component (Catlin's Automated Time Series Open Source) \cite{autoTS}.

Using our benchmarking framework shown in figure \ref{fig:bench}, we tested our system on a broad set of real world data sets and measured the performance of \emph{AutoAI-TS}'s various pipelines as well as the top performing pipelines against \emph{SOTA} frameworks. We tested our system on synthetic data described in section \ref{sec:syndata} for controlled experimentation. These experiments are performed to verify that our models capture characteristics based on model class and to check that models are appropriately ranked by the system given certain data characteristics and evaluated models' properties.

\subsection{Data Sets}
\label{sec:data}
\subsubsection{Synthetic Dataset}
\label{sec:syndata}
The synthetic data set contains total of \texttt{2000} data points and has \texttt{21} time series (total of \texttt{42,000} samples) that have different known signals such as linearly increasingly values, constants, linear increase with noise, exponential increase, inverse exponential, sine wave, cosine wave, sine and cosine wave with outliers, square wave function, sine and cosine signals with trend, log, exponential, wave form with dual seasonality etc.
The aim is to make sure our pipelines can capture such characteristics and are resilient to outliers in the data. Section \ref{sec:exp_syn} shows results of experiments with the synthetic data set.

\setlength{\tabcolsep}{7pt}
\begin{table}[t]
\scriptsize
  \centering
  \caption{List of Multivariate Time Series (Dimensions include the timestamp column)}
    \begin{tabular}{lcp{1.8in}}
    \hline
    Data set & \multicolumn{1}{l}{Size} & Source \\
    \hline
    walmart\_sale & (143, 11) & \textcolor[rgb]{ .02,  .388,  .757}{https://www.kaggle.com/c/walmart-recruiting-store-sales-forecasting} \\
    nn5tn10dim & (713, 11) &  \textcolor[rgb]{ .02,  .388,  .757}{http://www.neural-forecasting-competition.com}\\ 
    rossmann & (942, 11) & \textcolor[rgb]{ .02,  .388,  .757}{https://www.kaggle.com/c/rossmann-store-sales} \\
    household\_power & (1442, 10) & \textcolor[rgb]{ .02,  .388,  .757}{https://data.world/databeats/household-power-consumption} \\
    cloud & (2637, 5) & Proprietary source \\
    exchange\_rate & (7588, 9) & \cite{DBLP:journals/corr/LaiCYL17} \\
    traffic & (17544, 11) & \textcolor[rgb]{ .02,  .388,  .757}{http://pems.dot.ca.gov},\cite{DBLP:journals/corr/LaiCYL17} \\ 
    electricity & (26304, 11) & \textcolor[rgb]{ .02,  .388,  .757}{https://archive.ics.uci.edu/ml/datasets}, \cite{DBLP:journals/corr/LaiCYL17}  \\
    manufacturing & (303302, 6) & Propriety source \\
    \hline
    \end{tabular}%
  \label{tab:mts_list}%
\end{table}%

\subsubsection{Real World Dataset}
\label{sec:realdata}
We experimented with several real world data sets collected from different sources with a varying number of dimensions and records \cite{pyFTS:2018,tsdl:2018,DBLP:journals/corr/LaiCYL17,journals/ijon/AhmadLPA17}. 
%

\begin{figure*}[t]
\centering
\includegraphics[width=16cm]{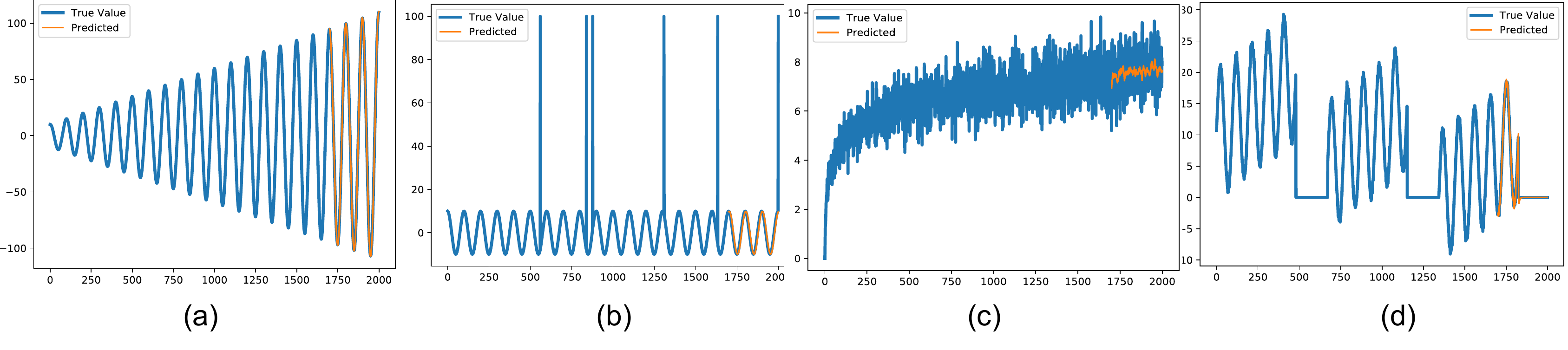}
\vspace*{-0.10cm}
\caption{Example synthetic signals: (a) cosine signal with increasing amplitude, (b) cosine signal with outliers, (c) logarithmic increase with variance, and (d) signal with multiple seasons.}
\vspace*{-0.2cm}
\label{fig:syn_results}
\end{figure*}

Using these data sets, we evaluate the system on \texttt{62} univariate time series (UTS) and \texttt{9} multivariate time series (MTS). The names and references to multivariate datasets are show in table \ref{tab:mts_list} and for the univariate datasets they are provided in appendix in table \ref{uts12-models-smape}. The number of data samples in the UTS vary from \texttt{144} to \texttt{145,366}, whereas in the MTS (with up to \texttt{10} dimensions) the number of samples vary between \texttt{143} and \texttt{303,302} (as reported in table~\ref{tab:mts_list}
). The list of UTS data sets is omitted from the paper due to space constraints. As we use a large number of time series for experimentation, many of the data sets do not have clean timestamps and some contain inconsistent timestamps. For such data sets, we re-generated their timestamps with day frequency, if they have less than \texttt{1000} data samples, otherwise their timestamps are generated with the \texttt{1} minute frequency. The data sets are collected from different domains such as social media (Twitter), finance, commerce, hydrology, agriculture, etc.

\subsection{Experiment 1: Synthetic Dataset}
\label{sec:exp_syn}
In this set of experiments, we evaluated the performance of our system on synthetic data, \texttt{1700} data samples are used for training and \texttt{300} for testing. The main purpose of this experiment was to verify that \emph{AutoAI-TS} pipelines can detect specific patterns of interest in the signals. In the interest of space, we show only \texttt{4} signals out of many different were tested.

Figure \ref{fig:syn_results}(a) shows \texttt{cosine} wave with increasing amplitude. As we can see from the figure showing output from one model that it can learn signal along with continuous increase in amplitude and predict signal accurately. Figure \ref{fig:syn_results}(b) shows a stable \texttt{cosine} wave injected with high peaking outliers spread unevenly through the training data. As seen from the figure that \emph{AutoAI-TS} can nicely model the main signal in the data predicting signal values and is resilient to outliers by not modeling the outliers. 

In figure \ref{fig:syn_results}(c), we tested our \emph{AutoAI-TS} models with a signal that follows a logarithmic increase with large variance around the mean. The goal is to test model resilience to high variance signals. As seen from the predicted values, our model is able to predict values and not responding too much to the variance. Figure \ref{fig:syn_results}(d) shows a signal that has dual seasonal patterns. As seen from predicted values, \emph{AutoAI-TS}
detects two seasons accurately and predicts values accordingly in respective seasons. For tests on synthetic data, error between actual and predicted value for all time series was below \texttt{1\%}.    

\subsection{Experiment 2: Real-World Datasets}

In this set of experiments, we evaluate our \emph{AutoAI-TS} against \texttt{10} SOTA models on real world data shown in section \ref{sec:realdata}. For all evaluated models, as well as individual pipelines within \emph{AutoAI-TS}, their hyper-parameters are kept as default and shown in table~\ref{tab:Parameter_setting}. We evaluate the performance of all models using Symmetric Mean Absolute Percentage Error (SMAPE) and the training time of models. It is important to note that, in line with our aim for \emph{zero-conf}, we did not optimize any parameters for running SOTA frameworks or \emph{AutoAI-TS} and evaluate performance of toolkits out-of-the-box without manual intervention or optimization. We believe that this approach presents a fair basis for comparison across toolkits from the non-expert end-user perspective. 
The results reported in this paper on SOTA frameworks are for comparison purposes under the mentioned assumptions and parameters. Readers interested in more in-depth benchmarking of these individual SOTA toolkits should consult their sources and respective published results.

\setlength\extrarowheight{2pt}
\begin{table*}[tbp]
\scriptsize
  \centering
   \caption{Parameter settings for all models in the experiments}
   \begin{tabular}{lp{70em}}
    \hline
    SOTA  & \multicolumn{1}{c}{Parameters} \\
    \hline
    DeepAR \cite{deeparpage} & num\_layers: 2, num\_cells: 40, cell\_type: lstm, dropout\_rate 0.1, use\_feat\_dynamic\_real False, use\_feat\_static\_cat False, use\_feat\_static\_real: False, distr\_output: StudentTOutput(), scaling: True, num\_parallel\_samples: 100 \\
    Prophet \cite{prophetgit} & n\_changepoints=25, changepoint\_range=0.8, yearly\_seasonality='auto', weekly\_seasonality='auto', daily\_seasonality='auto', holidays=None, seasonality\_mode='additive', mcmc\_samples=0, seasonality\_prior\_scale=10.0, holidays\_prior\_scale=10.0, changepoint\_prior\_scale=0.05, interval\_width=0.80, uncertainty\_samples=1000  \\
    Pmdarima \cite{pmdarima} & start\_p=1, start\_q=1, max\_p=3, max\_q=3, m=12, start\_P=0, seasonal=True, d=1, D=1\\
    Nbeats \cite{nbeatsgit}& thetas\_dims=[7, 8], nb\_blocks\_per\_stack=3,share\_weights\_in\_stack = False, train\_percent = 0.8, hidden\_layer\_units = 128  \\
    PyAF \cite{PyAF}  & mSignalDecomposition generated by tsdec.cSignalDecomposition() function, and mOptions generated by tsopts.cSignalDecomposition\_Options()  \\
    Component \cite{autoTS} & frequency: 'infer', prediction\_interval: 0.9, max\_generations: 1, no\_negatives: False, initial\_template: 'General+Random', random\_seed: 2020, holiday\_country: 'US', aggfunc: 'first', na\_tolerance: 1, metric\_weighting: {'smape\_weighting': 10, 'mae\_weighting': 2, 'rmse\_weighting': 2, 'containment\_weighting': 0, 'runtime\_weighting': 0, 'spl\_weighting': 1, 'contour\_weighting': 0}, drop\_most\_recent: 0, drop\_data\_older\_than\_periods: 100000, num\_validations: 2, models\_to\_validate: 0.15, validation\_method: 'even',min\_allowed\_train\_percent: 1.0, remove\_leading\_zeroes: False, model\_list: 'ComponentAnalysis'  \\
    GLS \cite{autoTS}  & Similar above except model\_list: 'GLS' \\
    WindowRegressor \cite{autoTS} & Similar above except model\_list: 'WindowRegressor' \\
    RollingRegressor \cite{autoTS} & Similar above except model\_list: 'RollingRegression'  \\
    Motif \cite{autoTS} & Similar above except model\_list: 'MotifSimulation' \\
    \textbf{AutoAI-TS} & 10 pipelines involved, 80\%-20\% as training-testing split for pipeline selection. No parameters tuning w.r.t each individual pipeline as well as the T-Daub \\
    \hline
    \end{tabular}%
  \label{tab:Parameter_setting}%
\end{table*}%

\begin{figure}[t]
\centering
\includegraphics[height=1.5in]{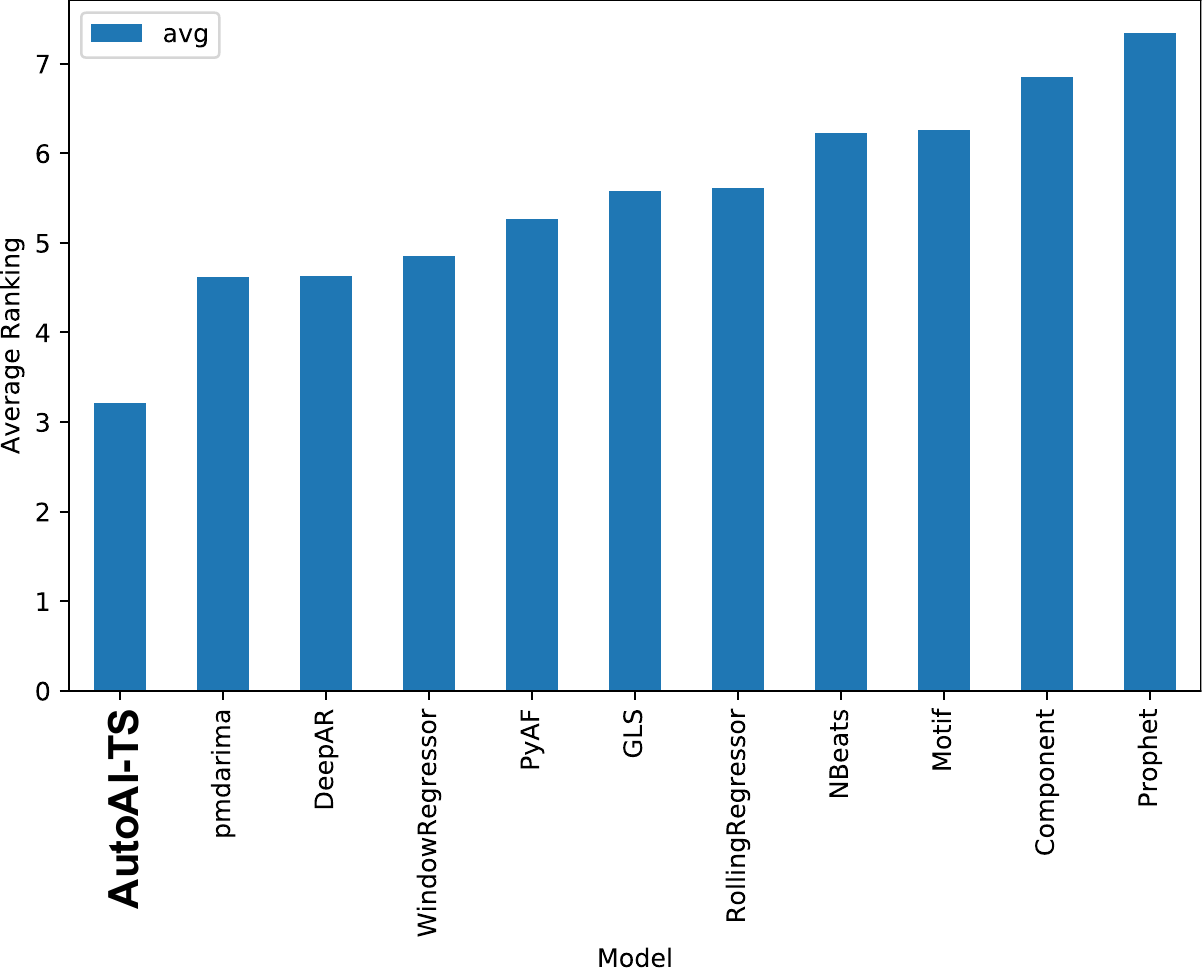}
\caption{SMAPE based comparison of \emph{AutoAI-TS} and SOTA toolkits for univariate data sets}
\label{fig:uni_smape}
\end{figure}

In our large-scale experiments, we vary the forecasting horizon between \texttt{6} and \texttt{30} in steps of \texttt{6}. The experimental results are quite consistent across these settings. Hence, we report results for the case where the prediction horizon is set to \texttt{12}, meaning models are trained to predict next \texttt{12} future values. For each individual time series, we rank the toolkits from \texttt{1} to \texttt{11} based on their SMAPE performance, with smaller ranks corresponding to low SMAPE values, namely higher performance (smaller is better). Figure~\ref{fig:uni_smape}, shows the average rank per model across all examined univariate timeseries. As seen from figure, the average rank of \emph{AutoAI-TS} is quite small indicating that its predicted values are more accurate comparing to the comparative toolkits on many time series. 

\begin{figure}[t]
\centering
\includegraphics[width=3.2in]{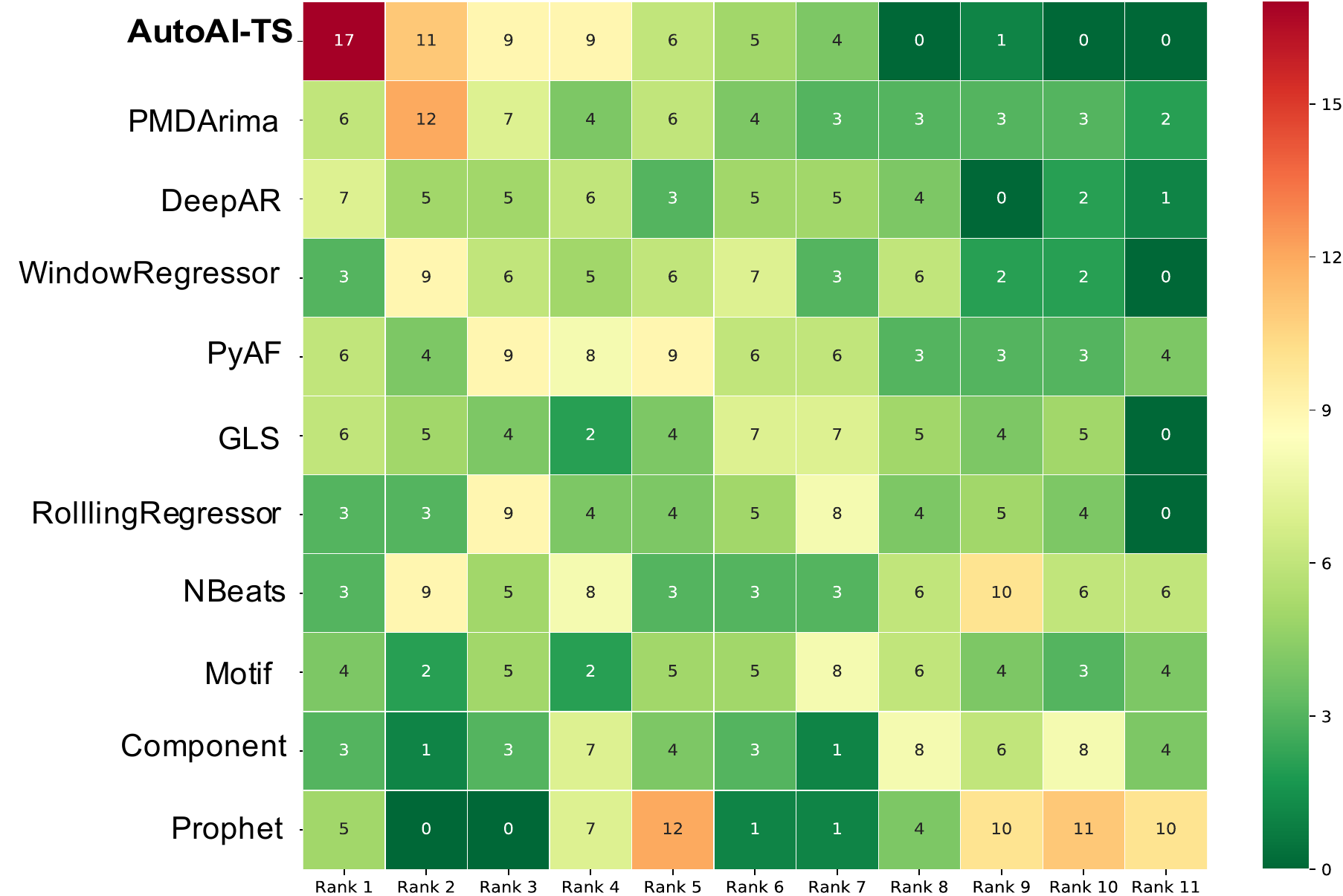}
\caption{SMAPE based ranking of \emph{AutoAI-TS} and SOTA toolkits for univariate data sets}
\label{fig:uni_smape_rank}
\vspace*{-0.50cm}
\end{figure}

Figure~\ref{fig:uni_smape_rank}, provides more details regarding the number of time series per rank per toolkit. One can observe that, out of \texttt{62} univariate time series, \emph{AutoAI-TS} achieves the best SMAPE performance on \texttt{17} time series, second best performance on \texttt{11} time series, and third and forth best performance on \texttt{9} time series each. \emph{Pmdarima} and \emph{DeepAR} perform quite well (2nd and 3rd respectively) and at the final rank is \emph{Prophet}. The low performance of \emph{Prophet} perhaps is due to the fact that our examined time series span across different domains, while \emph{Prophet} seems to work best for the business time series. It is also worth mentioning here that, not all models/toolkits can complete training on all \texttt{62} univariate time series within reasonable time and the above reported results are collected for those data sets for which experiments completed training and evaluation. We provide detailed accuracy scores and execution times of various toolkits for univariate experiments in table \ref{uts12-models-smape} in the appendix. 

\begin{figure}[t]
\centering
\includegraphics[height=1.5in]{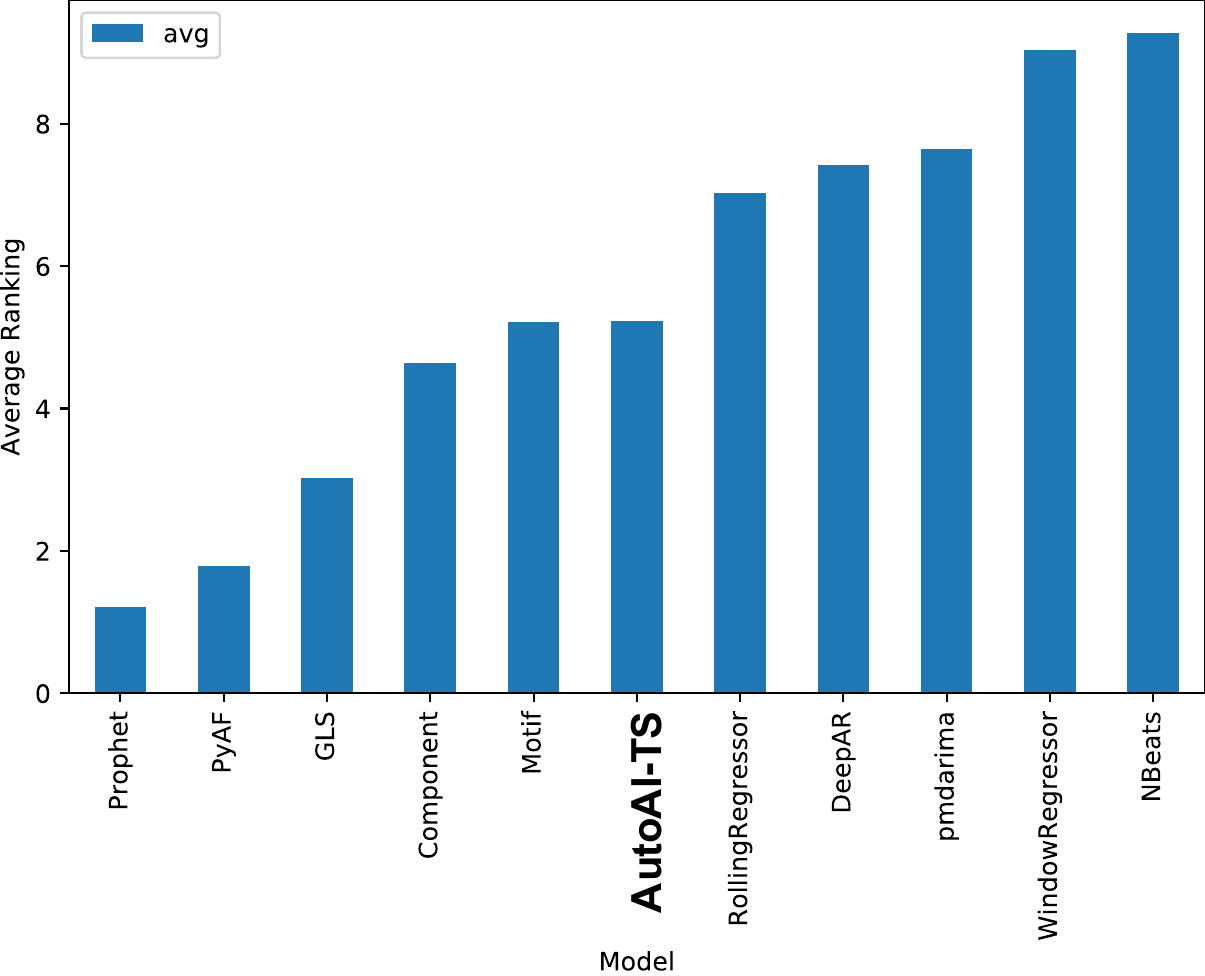}
\caption{Training time based comparison of \emph{AutoAI-TS} and SOTA toolkits for univariate data sets}
\label{fig:uni_traintime}
\end{figure}

\begin{figure}[t]
\centering
\includegraphics[width=3.2in]{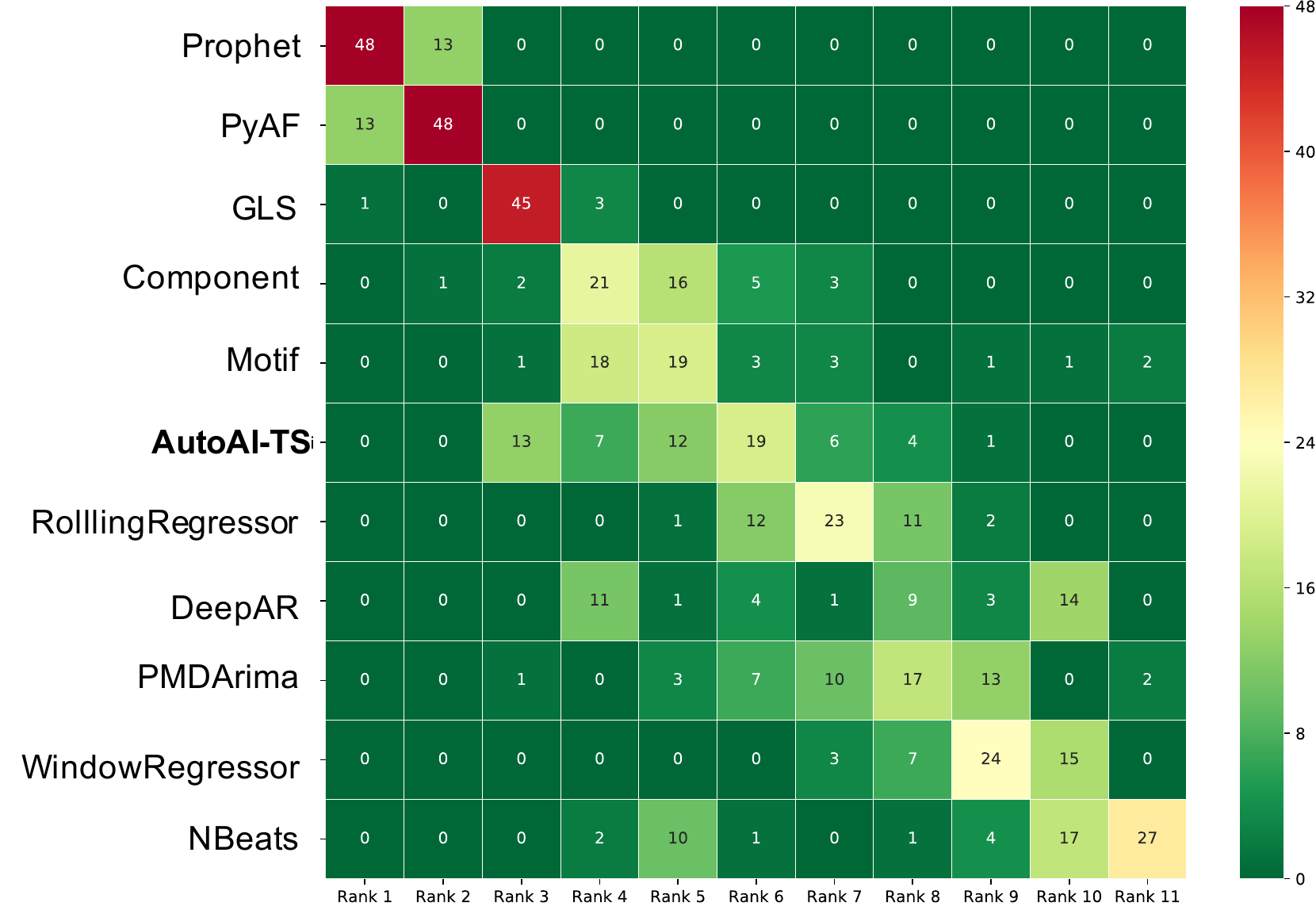}
\caption{Training time based ranking of \emph{AutoAI-TS} and SOTA toolkits for univariate data sets}
\label{fig:uni_traintime_rank}
\end{figure}

In terms of training time, in Figure~\ref{fig:uni_traintime} we plot the values averaged over all data sets. Similar to the forecasting accuracy, lower rank is better and lower rank means shorter training time for a given model/toolkit. For our \emph{AutoAI-TS} toolkit, the training time is the total time that T-Daub took until it discovered the best out of \texttt{10} pipelines (for a given time series) and retrained it on full data. 
\emph{AutoAI-TS} is slower than \emph{Prophet}, \emph{PyAF}, \emph{GLS}, \emph{Component} and \emph{Motif}, however, it is faster than \emph{RollingRegressor}, \emph{DeepAR}, \emph{Pmdarima}, \emph{WindowRegressor} and \emph{NBeats}. Long training time observed from \emph{DeepAR} and \emph{NBeats} are expected since they are deep learning models, whereas \emph{Pmdarima} often takes longer time on large time series. 

More details on the number of time series per rank per toolkit are provided in Figure~\ref{fig:uni_traintime_rank}. \emph{AutoAI-TS} is in the middle ranks in terms of training time, as it initially needs to train all \texttt{10} component pipelines using various allocations of the training data, and then fully re-trains the most promising pipeline with the highest potential to accurately model the given time series. Hence, \emph{AutoAI-TS} has majority of the data sets ranked between \texttt{3} and \texttt{6}, out of \texttt{11} toolkits.

\begin{figure}[t]
\centering
\includegraphics[height=1.5in]{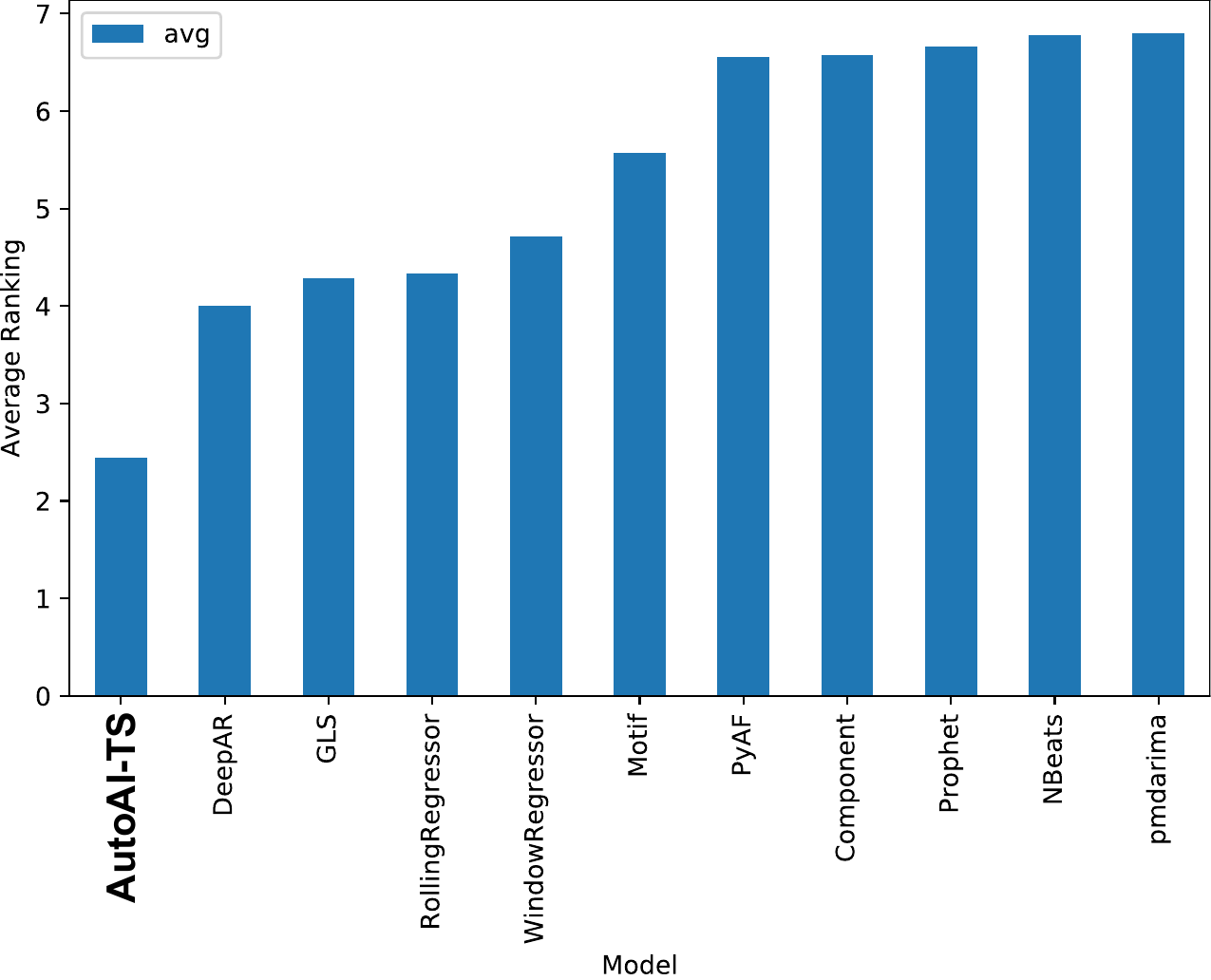}
\caption{SMAPE based comparison of \emph{AutoAI-TS} and SOTA toolkits for multivariate data sets}
\label{fig:mts_smape}
\end{figure}

\begin{figure}[t]
\centering
\includegraphics[width=3.2in]{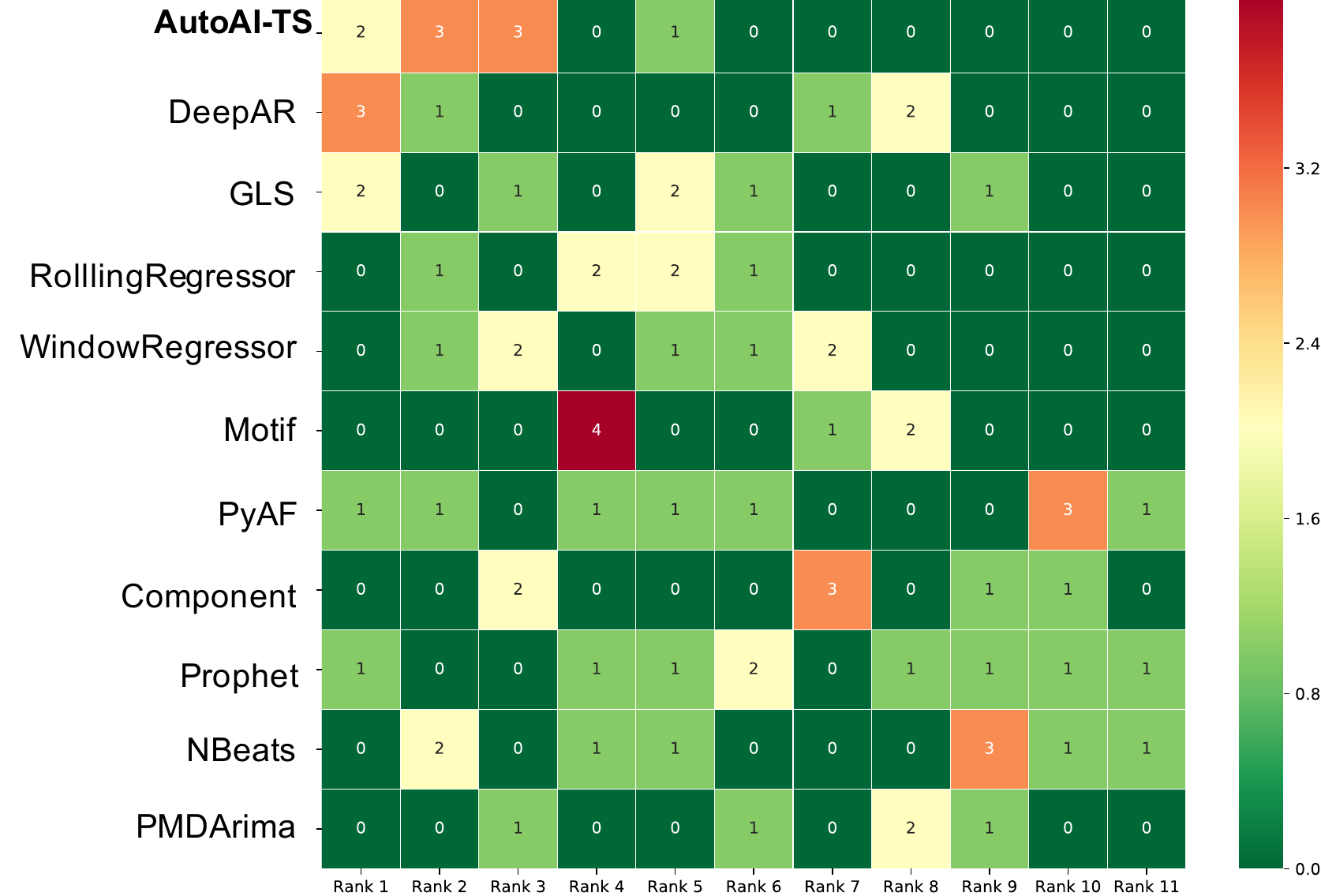}
\caption{SMAPE based ranking of \emph{AutoAI-TS} and SOTA toolkits for multivariate data sets}
\label{fig:mts_smape_rank}
\end{figure}

For prediction accuracy on multivariate data sets (MTS), Figure~\ref{fig:mts_smape} shows the average rank across all nine data sets for all models. \emph{AutoAI-TS} performance remains consistently good, on average, and it outperforms other SOTA toolkits. As further shown in Figure~\ref{fig:mts_smape_rank}, \emph{DeepAR} also performs well with best SMAPE on \texttt{3} out of \texttt{9} data sets, while \emph{Pmdarima} seems to be not performing as good as other toolkits. \emph{AutoAI-TS} achieves best SMAPE on \texttt{2} data sets and remains \texttt{2nd} and \texttt{3rd} best SMAPE on other six data sets out of nine. The detailed scores and execution times are provided in table \ref{tab:mts12-models-smape} in the appendix. 

In summary, we believe that the good performance accuracy of \emph{AutoAI-TS} is due to the fact that it encompasses a  diverse set of techniques, including statistical models, exponential smoothing, machine learning-based ones, etc. This key feature allows \emph{AutoAI-TS} to choose a suitable technique and pipeline for a given time series regardless of the domain of a data set. At the same time, T-Daub assists in quickly choosing the right pipeline for full data training. It is important to note that \emph{AutoAI-TS} is very competitive even when compared to other toolkits that incorporate newer, state-of-the-art deep learning models. 

\begin{figure}[t]
\centering
\includegraphics[height=1.5in]{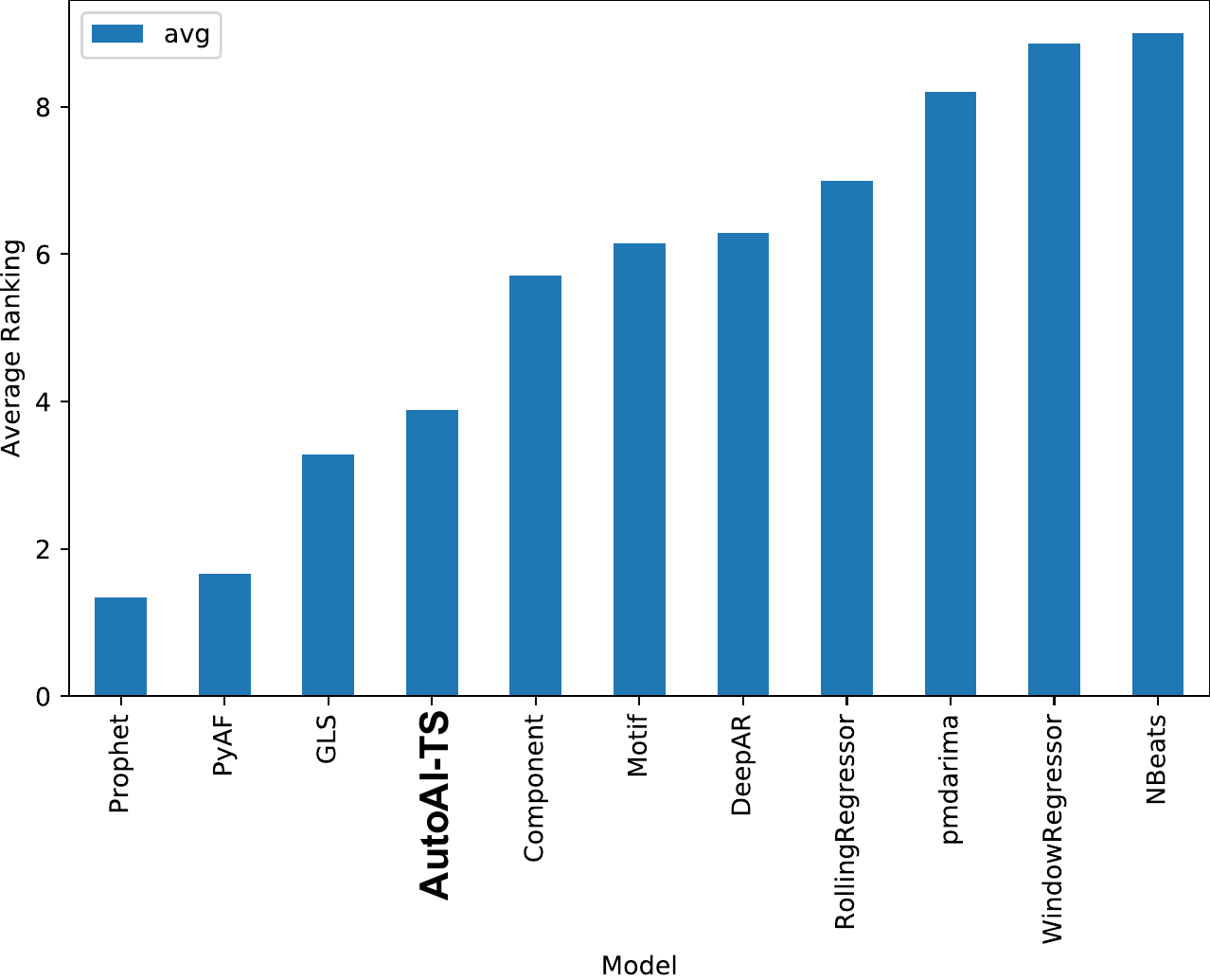}
\caption{Training time based comparison of \emph{AutoAI-TS} and SOTA toolkits for multivariate data sets}
\label{fig:mts_traintime}
\end{figure}

\begin{figure}[t]
\centering
\includegraphics[width=3.4in]{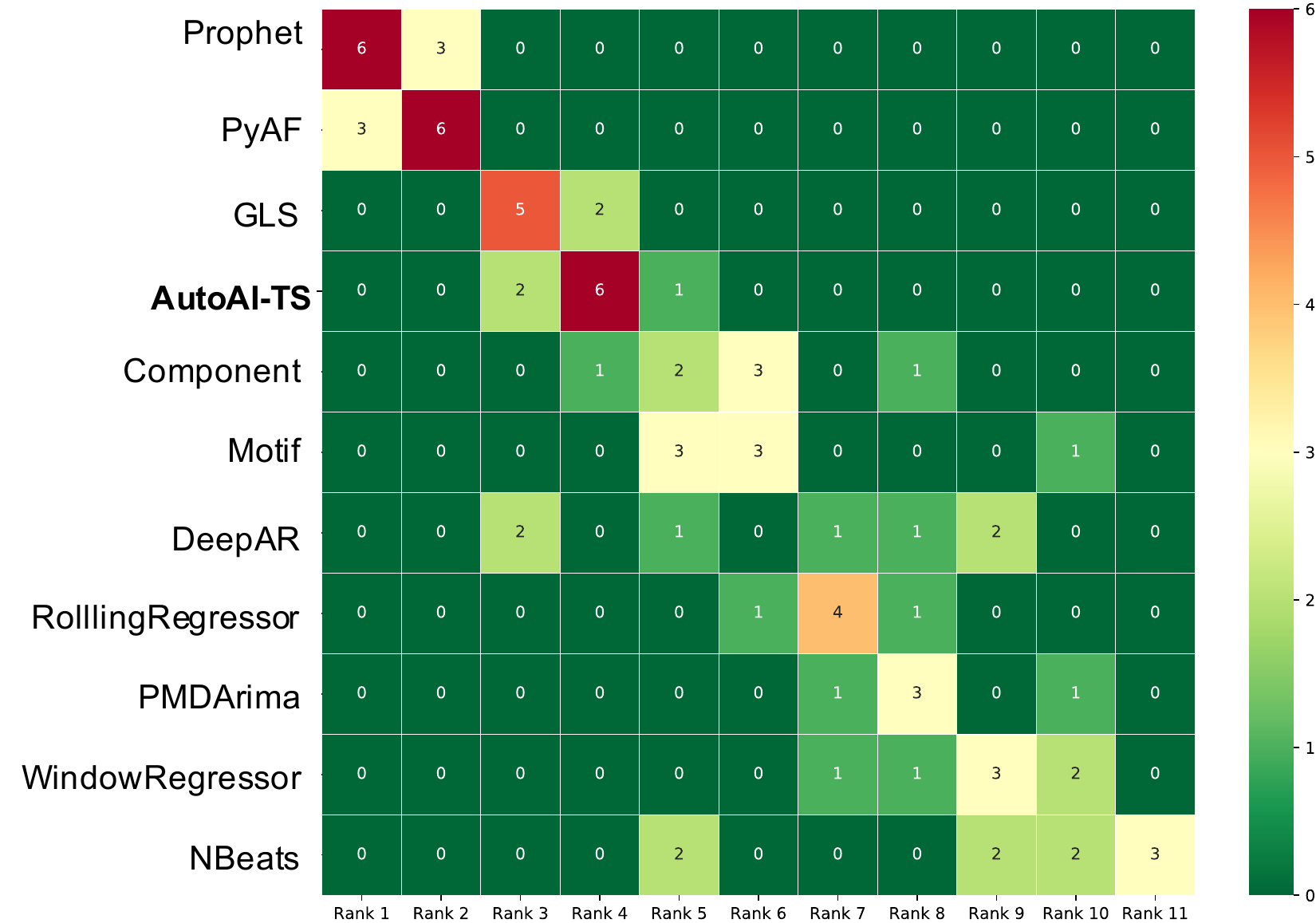}
\caption{Training time based ranking of \emph{AutoAI-TS} and SOTA toolkits for multivariate data sets}
\label{fig:mts_traintime_rank}
\end{figure}

Figure~\ref{fig:mts_traintime} illustrates the ranking for training time for all models averaged over \texttt{9} MTS data sets. Figure~\ref{fig:mts_traintime_rank} shows in detail the number of time series per rank per model. \emph{AutoAI-TS} similarly ranks in the middle in terms of training time and compares favorably to other SOTA toolkits such as Component, \emph{DeepAR}, and others, while retaining good forecasting accuracy performance. These results are observed from \texttt{9} typical multivariate time series. Since this is a relatively small number, we plan to extend these experiments in the near future to include more multivariate data sets. Moreover, as mentioned before, for SOTA techniques no manual optimization was performed, which might make them run faster at the cost of prediction accuracy. 

\begin{figure}[t]
\centering
\includegraphics[height=2.0in]{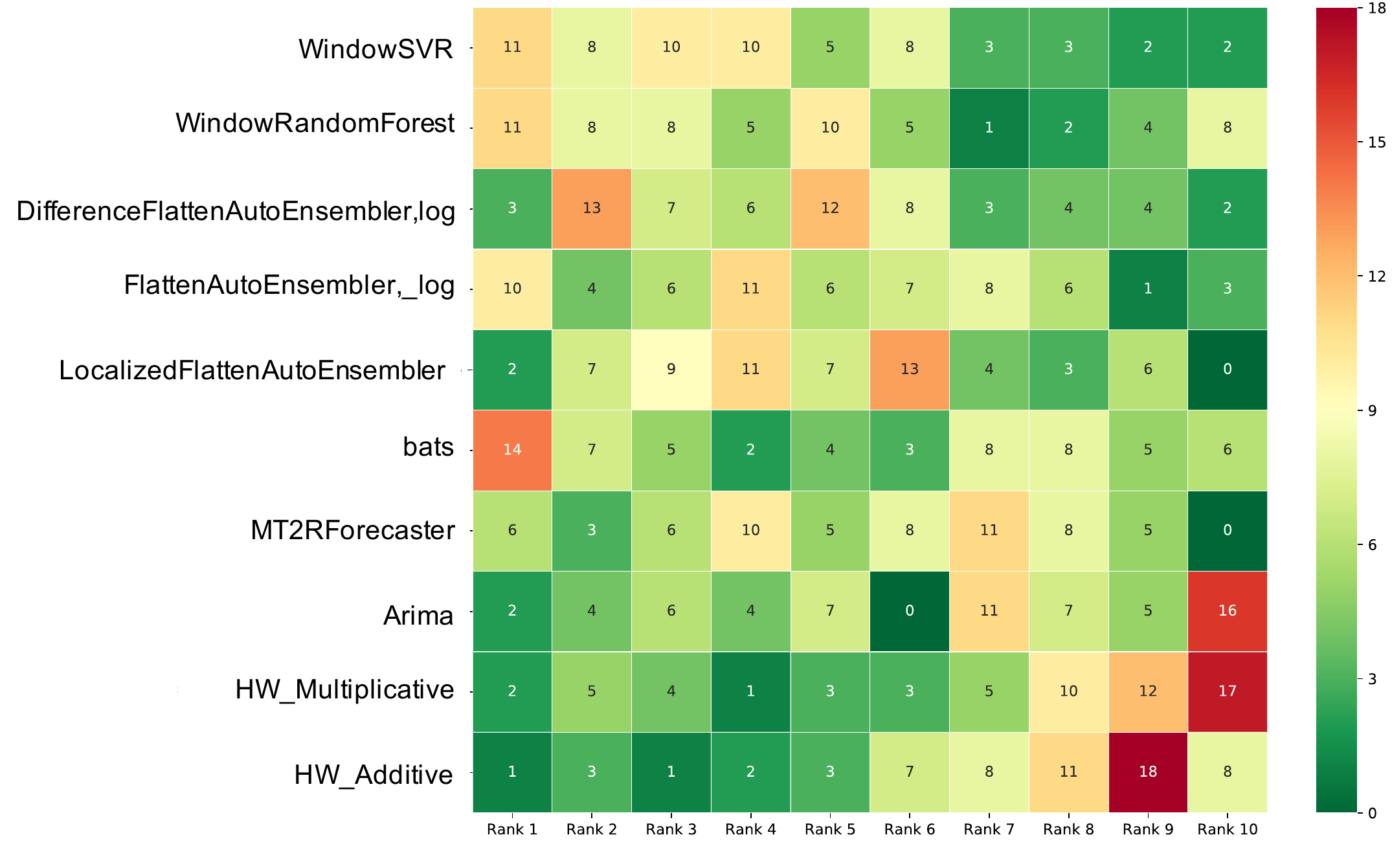}
\caption{Ranking of \emph{AutoAI-TS} pipelines on univariate benchmarked data sets}
\label{fig:uts_inter_rank}
\end{figure}

\begin{figure}[t]
\centering
\includegraphics[height=2.2in]{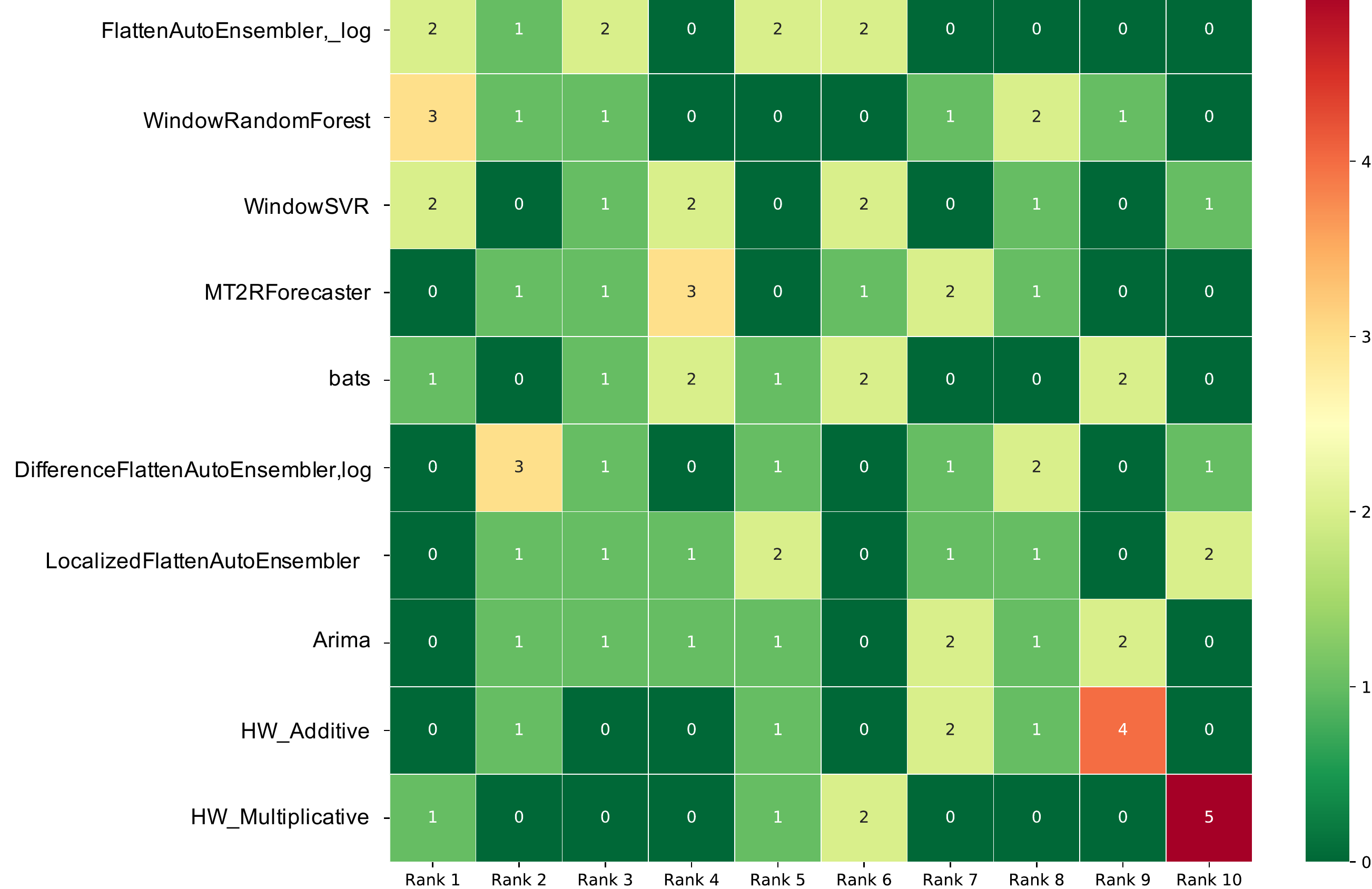}
\caption{Ranking of \emph{AutoAI-TS} pipelines on multivariate benchmarked data sets}
\label{fig:mts_inter_rank}
\end{figure}

Figure \ref{fig:uts_inter_rank} shows the ranking of the \texttt{10} test pipelines of \emph{AutoAI-TS} on univariate data sets. As we can see from the figure, no single model works best on all \texttt{62} data sets; in fact, the top \texttt{3} ranks have a spread of various models, which validates our hypothesis for having models from different model classes. Similarly, Figure \ref{fig:mts_inter_rank} shows the ranking of various \emph{AutoAI-TS} pipelines on multivariate data sets. Since we have only \texttt{9} multivariate data sets which is relatively low number still we see that more than one models are ranked in top \texttt{3} spots. From Figures \ref{fig:uts_inter_rank} and \ref{fig:mts_traintime_rank} we conclude that having multiple models  helps \emph{AutoAI-TS} cover many data sets from diverse domains and achieve good forecasting accuracy. We show detailed scores and execution times of internal pipelines of \emph{AutoAI-TS} in table \ref{tab:mts12-ppln-smape} only for multivariate data sets due space constraints.

\emph{AutoAI-TS} tries to perform better on average on various datasets without pre-selecting pipelines. It tries to achieve trade-off between achieving accuracy and compute time, therefore it does not always performs the best on all benchmarks against SOTA. There are cases where we believe the AutoAI-TS might perform poor in terms of accuracy or compute time. For example, if the training data used to select pipeline has different characteristics than then test data \emph{AutoAI-TS} might have poor accuracy because \emph{AutoAI-TS} chooses pipeline based on training data. In terms of execution time, \emph{AutoAI-TS} might perform poor in cases where there is well known season or trend in the data which simple algorithms like ARIMA can model quickly, the \emph{AutoAI-TS} might consume more time in evaluating all the pipelines before closing best model. Nevertheless, as it selects among a number of algorithms we are generally able to eliminate algorithms which perform poorly and hence we still can achieve good performance. 

\section{Conclusion}
\label{sec:conclusion}
This paper described the design, implementation and performance evaluation of \emph{AutoAI-TS}, a system framework for the automation of time series forecasting task. \emph{AutoAI-TS} transparently integrates several classes of predictive models and enables immediate use by non-expert as well as expert users with zero-conf. \emph{AutoAI-TS} automatically handles feature engineering, model training and pipeline ranking and selects best models for deployment and inference. In addition to a variety of time series modeling techniques, \emph{AutoAI-TS} incorporates an automatic pipeline ranking and selection mechanism to quickly identify the most promising pipelines for a given dataset. In future, we plan to extend AutoAI-TS in various directions such as adding anomaly detection, high volatility models, causal analysis of time series and various kinds of deep learning models.




\section{Appendix}
\setlength\extrarowheight{4pt}
\begin{table*}[htbp]
  \centering
\caption{Symmetric mean absolute percentage error (smape) of all models along with execution time in seconds in parenthesis on \texttt{62} univariate time series. The values displayed in each columns are "smape (total execution in seconds)". Smaller smape values indicate better performance. 0(0) value indicates a model did not finish in reasonable time w.r.t other models on the corresponding dataset and it was terminated. The datasets are ordered based on their sizes ranging from \texttt{144} data observations (AirPassengers) to \texttt{145,366} data observations (PJME-MW). Models are ordered from left to right according to their average performance shown in Fig.\ref{fig:uni_smape_rank}.}
 \resizebox{\textwidth}{!}{%

 \begin{tabular}{rlrrrrrrrrrrr}
    \toprule
    Index & Datasets & \multicolumn{1}{r}{AutoAI-TS} & \multicolumn{1}{r}{PMDArima} & \multicolumn{1}{r}{DeepAR} & \multicolumn{1}{r}{WindowRegressor} & \multicolumn{1}{r}{PyAF} & \multicolumn{1}{r}{GLS} & \multicolumn{1}{r}{RollingRegressor} & \multicolumn{1}{r}{NBeats} & \multicolumn{1}{r}{Motif} & \multicolumn{1}{r}{Component} & \multicolumn{1}{r}{Prophet} \\
    \midrule
    1     & AirPassengers \cite{pyFTS:2018} & 2.7 (50.75) & 4.62 (4.03) & 6.29 (267.31) & 5.32 (46.91) & 3.56 (3.39) & 10.02 (17.34) & 5.28 (48.95) & 6.85 (383.3) & 18.78 (19.86) & 14.55 (22.87) & 12.23 (2.03) \\
    2     & a10 \cite{dataset1:2018}  & 8.87 (49.36) & 7.22 (28.17) & 10.83 (265.04) & 7.64 (57.04) & 8.55 (3.7) & 12.28 (17.86) & 9.77 (40.71) & 21.22 (409.53) & 11.65 (18.1) & 19.53 (24.89) & 14.17 (2.05) \\
    3     & h02 \cite{dataset1:2018}  & 6.24 (48.81) & 6.74 (205.42) & 9.43 (264.52) & 7.41 (64.44) & 5.16 (3.69) & 8.76 (18.06) & 6.77 (40.36) & 6.82 (400.42) & 20.51 (18.53) & 14.45 (20.26) & 20.04 (1.98) \\
    4     & ausbeer \cite{dataset1:2018} & 1.53 (50.51) & 1.55 (46.34) & 1.65 (303.24) & 2.91 (68.29) & 2.44 (3.85) & 8.08 (17.79) & 5.64 (44.14) & 6.37 (410.37) & 2.94 (20.81) & 9.02 (15.15) & 7.02 (2.33) \\
    5     & qauselec \cite{dataset1:2018} & 2.78 (51.62) & 3.94 (196.64) & 2.23 (301.87) & 2.7 (75) & 3.22 (3.72) & 2.77 (15.99) & 3.01 (36.25) & 4.98 (407.12) & 4.47 (20.78) & 3.15 (18.51) & 3.34 (2.44) \\
    6     & qgas \cite{dataset1:2018} & 1.71 (48.19) & 2.29 (37.08) & 4.3 (302.44) & 3.23 (74.53) & 3.71 (3.72) & 9.66 (18.11) & 9.93 (35.22) & 14.82 (420.48) & 9.79 (20.02) & 6.1 (20.6) & 9.91 (2.4) \\
    7     & ozone \cite{dataset1:2018} & 15.52 (50.84) & 28.48 (45.83) & 0 (0) & 19.19 (60.98) & 22.76 (3.7) & 31.17 (17.98) & 19.7 (54.54) & 35.98 (402.93) & 31.37 (18.18) & 33.22 (31.81) & 40.82 (2.47) \\
    8     & qcement \cite{dataset1:2018} & 4.59 (50.19) & 6.78 (23.03) & 9.73 (302.33) & 4.37 (66.94) & 10.53 (7.25) & 7.65 (17.42) & 11.37 (63.49) & 16.56 (434.03) & 6.1 (22.01) & 7.92 (21.92) & 6.69 (2.33) \\
    9     & melsyd \cite{dataset1:2018} & 9.76 (49.95) & 18.47 (51.62) & 42.13 (307.13) & 24.75 (81.96) & 33.55 (7.32) & 7.43 (18.56) & 12.87 (51.03) & 9.42 (485.86) & 32.36 (21.66) & 57.66 (22.49) & 13.23 (2.36) \\
    10    & elecdaily \cite{dataset1:2018} & 8.88 (50.74) & 10.31 (157.1) & 10.41 (301.79) & 11.4 (103.55) & 13.33 (8.97) & 11.97 (17.78) & 11.18 (53.14) & 11.33 (529.93) & 10.88 (28.83) & 8.92 (40.52) & 8.63 (2.14) \\
    11    & hyndsight \cite{dataset1:2018} & 8.8 (55.56) & 16.1 (67.07) & 9.3 (306.7) & 7.6 (95.22) & 8.28 (7.28) & 16.18 (18.15) & 6.76 (49.12) & 36.3 (501) & 11.98 (22.67) & 8.77 (48.36) & 8.41 (2.44) \\
    12    & Births \cite{dataset1:2018} & 13.63 (50.3) & 14.96 (60.32) & 15.28 (311.32) & 13.6 (127.53) & 13.73 (7.59) & 13.53 (18.51) & 13.56 (51.16) & 16.18 (505.3) & 13.95 (25.45) & 15.27 (21.79) & 13.14 (2.5) \\
    13    & auscafe \cite{dataset1:2018} & 2.03 (51.57) & 1.35 (49.33) & 5.6 (315.01) & 1.63 (147.93) & 1.89 (8.08) & 4.25 (17.61) & 6.79 (45.54) & 11.68 (704.65) & 4.42 (27.44) & 12.92 (25.3) & 3.96 (2.48) \\
    14    & usmelec \cite{dataset1:2018} & 1.8 (52.56) & 1.45 (54.59) & 2.53 (315.47) & 3.06 (151.61) & 2.72 (7.68) & 8.43 (16.84) & 4.52 (69.37) & 2.47 (719.53) & 11.62 (25.03) & 8.28 (28.74) & 9.85 (2.47) \\
    15    & departures \cite{dataset1:2018} & 11.08 (51.92) & 10.45 (37.2) & 9.99 (320.76) & 10.54 (182.29) & 11.95 (7.77) & 16.03 (15.58) & 17.81 (69.02) & 11.77 (717.16) & 16.73 (31.96) & 18.53 (22.49) & 17.54 (2.5) \\
    16    & goog \cite{dataset1:2018}  & 1.52 (65.73) & 1.57 (140.58) & 0 (0) & 2.45 (220.02) & 0 (0) & 1.57 (17.86) & 2.79 (68.96) & 10.26 (827.61) & 1.79 (38.6) & 1.68 (34.09) & 0 (0) \\
    17    & speed \cite{dataset1:2018} & 45.1 (70.15) & 43.05 (176.74) & 0 (0) & 48.24 (376.29) & 39.16 (8.52) & 47.95 (17.42) & 48.06 (116.29) & 30.88 (884.79) & 52.13 (54.38) & 51.5 (21.7) & 23.8 (2.67) \\
    18    & gasoline \cite{dataset1:2018} & 2.25 (68.88) & 2.82 (218.55) & 1.98 (342.3) & 2.23 (358.3) & 3.14 (8.88) & 5.96 (17.48) & 3.83 (116.7) & 2.51 (957.52) & 3.12 (48.3) & 5.17 (73.67) & 90.47 (2.79) \\
    19    & exchange-3-cpc-results \cite{journals/ijon/AhmadLPA17} & 21.8 (90.29) & 30.64 (328.83) & 0 (0) & 38.28 (744.91) & 33.11 (8.08) & 37.31 (16.36) & 38.68 (170.57) & 26.26 (1000.67) & 32.14 (48.58) & 52.13 (54.48) & 30.86 (2.67) \\
    20    & exchange-3-cpm-results \cite{journals/ijon/AhmadLPA17} & 18.65 (95.75) & 33.51 (258.21) & 0 (0) & 36.7 (532.14) & 22.81 (8.58) & 35.37 (20.57) & 35.51 (134.72) & 28.71 (1006.09) & 35.56 (148.8) & 32.19 (15.08) & 31.36 (3.02) \\
    21    & exchange-2-cpc-results \cite{journals/ijon/AhmadLPA17} & 15.51 (92.05) & 14.2 (1664.32) & 21.32 (360.11) & 16.7 (510.39) & 18.71 (8.23) & 103.77 (18.36) & 19.06 (167.75) & 24.08 (1011.43) & 103.36 (56.38) & 16.16 (233.88) & 156.5 (2.68) \\
    22    & exchange-2-cpm-results \cite{journals/ijon/AhmadLPA17} & 13.28 (97.96) & 18.16 (1273.09) & 33.46 (354.79) & 67.13 (572.7) & 19.3 (8.33) & 29.43 (18.3) & 18.78 (148.92) & 32.3 (1018.82) & 39.64 (81.16) & 49.05 (28.08) & 76.77 (2.48) \\
    23    & exchange-4-cpc-results \cite{journals/ijon/AhmadLPA17} & 90.02 (92.94) & 63.52 (334.03) & 0 (0) & 62.97 (449.98) & 63.2 (8.32) & 60.86 (19.86) & 145.49 (341.02) & 51.79 (1003.81) & 74.1 (60.52) & 82.35 (40.15) & 80 (2.91) \\
    24    & exchange-4-cpm-results \cite{journals/ijon/AhmadLPA17} & 51.31 (92.78) & 77.65 (311.13) & 0 (0) & 79.1 (478.82) & 81.68 (8.3) & 74.49 (19.41) & 54.37 (100.52) & 53.74 (1026.13) & 85.7 (62.33) & 73.77 (30.23) & 94.22 (2.53) \\
    25    & TravelTime-451 \cite{journals/ijon/AhmadLPA17} & 40.85 (102.95) & 14.94 (334.33) & 0 (0) & 48.74 (698.76) & 32.45 (8.63) & 34 (17.64) & 24.65 (204.05) & 48.76 (1158.47) & 30.92 (51.17) & 11.27 (74.64) & 169.07 (3.18) \\
    26    & occupancy-6005 \cite{journals/ijon/AhmadLPA17} & 42.99 (136.98) & 45.41 (298.95) & 0 (0) & 42.17 (988.43) & 200 (8.91) & 44.32 (19.4) & 54.74 (184.67) & 43.25 (1198.43) & 39.43 (119.93) & 37.22 (34.55) & 35.22 (2.88) \\
    27    & speed-t4013  \cite{journals/ijon/AhmadLPA17} & 4.14 (123.98) & 6.08 (332.5) & 3.86 (409.89) & 3.95 (623.9) & 4.39 (8.74) & 3.7 (22.36) & 3.65 (214.95) & 4.39 (2574.86) & 3.48 (506.49) & 4.08 (75.25) & 133.52 (2.61) \\
    28    & TravelTime-387 \cite{journals/ijon/AhmadLPA17} & 25.6 (122.21) & 41.12 (408.11) & 0 (0) & 85.87 (1139.56) & 27.84 (9.55) & 52.07 (19.15) & 26.78 (288.72) & 27.34 (2571.81) & 16.47 (63.07) & 85.05 (28.07) & 120.34 (4.05) \\
    29    & occupancy-t4013 \cite{journals/ijon/AhmadLPA17} & 22.24 (123.48) & 20.08 (393.85) & 30.9 (399.75) & 19.97 (860.08) & 200 (8.3) & 20.46 (20.14) & 28.59 (215.95) & 56.84 (2576.39) & 21.88 (153.85) & 19.96 (254.01) & 111.43 (3) \\
    30    & speed-6005 \cite{journals/ijon/AhmadLPA17} & 3.68 (122.33) & 4.31 (390.48) & 0 (0) & 3.68 (919.14) & 4.37 (8.85) & 4.01 (18.18) & 4.05 (217.87) & 5.98 (2582.93) & 3.92 (60.08) & 3.94 (99.78) & 4.58 (2.93) \\
    31    & Sunspots \cite{pyFTS:2018} & 21.71 (189.15) & 31.19 (441.76) & 27.15 (444.28) & 22.34 (1058.87) & 200 (9.69) & 46.63 (19.23) & 57.95 (303.09) & 48.5 (2680.99) & 111.79 (170.32) & 67.55 (76.84) & 96.93 (3.67) \\
    32    & Min-Temp \cite{pyFTS:2018} & 8.39 (135.46) & 8.57 (596.14) & 0 (0) & 14.17 (1436.31) & 10.36 (10.26) & 6.61 (20.51) & 10.43 (281.72) & 29.3 (2737.56) & 12.37 (166.43) & 104.81 (132.81) & 9.77 (3.23) \\
    33    & ec2-cpu-utilization-24ae8d \cite{journals/ijon/AhmadLPA17} & 6.91 (214.27) & 16.16 (744.02) & 8.98 (474.32) & 3.58 (1046.49) & 0.88 (9.82) & 0.5 (17.72) & 9.51 (256.91) & 9.18 (3420.21) & 0.5 (282.85) & 13.22 (237.97) & 5.67 (2.97) \\
    34    & ec2-cpu-utilization-53ea38 \cite{journals/ijon/AhmadLPA17} & 2.06 (206.56) & 1.89 (697.7) & 2.04 (473.47) & 2.38 (1531.24) & 2.2 (9.52) & 3.56 (20.43) & 1.82 (655.15) & 2.42 (3429.46) & 3.97 (117.88) & 3.82 (125.72) & 3.37 (2.81) \\
    35    & ec2-cpu-utilization-5f5533 \cite{journals/ijon/AhmadLPA17} & 1.79 (205.42) & 1.78 (697.24) & 1.71 (479.79) & 1.98 (2966.51) & 1.8 (9.72) & 2.28 (20.2) & 2.25 (395.93) & 2.33 (3433.34) & 2.08 (155.48) & 2.46 (103.74) & 2.82 (3.76) \\
    36    & ec2-cpu-utilization-77c1ca \cite{journals/ijon/AhmadLPA17} & 143.54 (209.95) & 194.45 (482.68) & 89.79 (487.86) & 89.6 (1420.05) & 14.31 (10.89) & 9.48 (24.49) & 144.16 (702.66) & 190.5 (3485.78) & 9.48 (150.39) & 45.43 (52.25) & 197.5 (3.44) \\
    37    & ec2-cpu-utilization-825cc2 \cite{journals/ijon/AhmadLPA17} & 1.16 (191.22) & 1.35 (323.6) & 0 (0) & 1.07 (8337.59) & 4.31 (9.88) & 0.95 (23.01) & 1.05 (329.69) & 1.15 (3440.18) & 1.3 (178.47) & 0.94 (202.87) & 8.89 (6.62) \\
    38    & ec2-cpu-utilization-ac20cd \cite{journals/ijon/AhmadLPA17} & 0.46 (184.51) & 0.39 (410.07) & 0 (0) & 0.5 (4889.31) & 0.64 (9.38) & 0.55 (21.63) & 0.41 (638.37) & 26.74 (3444.77) & 0.45 (160.72) & 0.78 (173.73) & 97.28 (4.97) \\
    39    & ec2-cpu-utilization-c6585a \cite{journals/ijon/AhmadLPA17} & 17.48 (177.16) & 21.96 (842.85) & 39.1 (474.08) & 19.38 (1839.78) & 18 (9.45) & 10.48 (22.88) & 22.41 (536.26) & 26.66 (3428.02) & 17.55 (486.89) & 18 (127.46) & 23.76 (2.39) \\
    40    & ec2-cpu-utilization-fe7f93 \cite{journals/ijon/AhmadLPA17} & 32.43 (182.38) & 35.26 (528.24) & 5.85 (473.99) & 32.64 (1440.12) & 29.18 (10.1) & 12.09 (22.21) & 13.95 (393.21) & 36.58 (3413.71) & 17.09 (255.18) & 20.95 (164.69) & 109.02 (3.52) \\
    41    & ec2-network-in-257a54 \cite{journals/ijon/AhmadLPA17} & 6.07 (222.96) & 4.39 (260.19) & 0 (0) & 4.9 (1571.64) & 7.06 (10.17) & 4.12 (19.62) & 6.08 (398.04) & 120.04 (3410.46) & 7.69 (205.47) & 0 (0) & 24.33 (4.15) \\
    42    & elb-request-count-8c0756 \cite{journals/ijon/AhmadLPA17} & 112.62 (179.28) & 116 (708.62) & 0 (0) & 99.8 (1544.49) & 200 (9.37) & 105.96 (43.77) & 119.84 (449.47) & 120.38 (3433.39) & 91.25 (242.38) & 132.73 (101.8) & 140.72 (3.55) \\
    43    & rds-cpu-utilization \cite{journals/ijon/AhmadLPA17} & 4.86 (197.35) & 4.68 (611.39) & 0 (0) & 4.77 (12317.63) & 5.08 (12.38) & 5.44 (20.68) & 3.7 (300.2) & 4.83 (3414.84) & 5.3 (86.27) & 8.15 (62.95) & 10.28 (6.54) \\
    44    & rds-cpu-utilization \cite{journals/ijon/AhmadLPA17} & 1.82 (228.32) & 1.62 (625.56) & 1.79 (480.34) & 2.2 (1262.72) & 2.44 (9.96) & 4.1 (20.29) & 2.08 (267.73) & 1.9 (3746.36) & 3.2 (238.78) & 6.85 (323.81) & 182.57 (7.37) \\
    45    & ec2-network-in-5abac7 \cite{journals/ijon/AhmadLPA17} & 39.23 (255.47) & 199.34 (310.46) & 3.1 (482.01) & 24.75 (1514.27) & 45.17 (9.27) & 51.26 (20.83) & 24.06 (488.37) & 197.75 (3458.09) & 37.89 (241.73) & 30.34 (302.25) & 200 (3.38) \\
    46    & Twitter-volume-AMZN \cite{journals/ijon/AhmadLPA17} & 17.07 (562.32) & 24.37 (2534.27) & 20.19 (1049.26) & 20.73 (2348.88) & 21.17 (16.16) & 17.88 (45.68) & 18.54 (1660.7) & 24.33 (3317.82) & 23.95 (3311.55) & 18.73 (926.03) & 23.91 (14.08) \\
    47    & Twitter-volume \cite{journals/ijon/AhmadLPA17} & 62.25 (490.21) & 38.45 (2280.29) & 80.95 (1044.89) & 95.45 (1628.46) & 200 (16.99) & 62.65 (35.91) & 123.1 (1900.07) & 33.56 (3348.13) & 77.3 (4246.54) & 114.6 (437.9) & 129.52 (13.87) \\
    48    & Twitter-volume-GOOG \cite{journals/ijon/AhmadLPA17} & 16.57 (402.53) & 17.18 (3054.45) & 28.81 (1049.68) & 12.86 (2332.3) & 200 (15.75) & 21.74 (33.55) & 17.54 (2025.88) & 31.5 (3364.92) & 0 (0) & 18.56 (750.26) & 40.58 (20.95) \\
    49    & Twitter-volume-AAPL \cite{journals/ijon/AhmadLPA17} & 25.26 (648.89) & 29.58 (2411.35) & 20.02 (1051.17) & 33.34 (2151.67) & 21.08 (15.61) & 37.48 (58.1) & 28.94 (1217.31) & 53.76 (3375.66) & 20.8 (3410.62) & 21.03 (1105.08) & 63.12 (9.3) \\
    50    & elecdemand \cite{dataset1:2018} & 4.11 (597.43) & 10.02 (2746.85) & 12.24 (1039.38) & 0 (0) & 7.46 (13.74) & 0 (0) & 0 (0) & 3.15 (3430.92) & 0 (0) & 0 (0) & 1.91 (28.28) \\
    51    & calls \cite{dataset1:2018} & 7.66 (792.87) & 9.44 (3943.26) & 7.96 (1544.01) & 0 (0) & 20.5 (19.01) & 0 (0) & 0 (0) & 8.62 (3875.76) & 0 (0) & 0 (0) & 91.12 (12.58) \\
    52    & PJM-Load-MW \cite{dataset2:2018} & 12.62 (848.27) & 2.59 (10012.16) & 0 (0) & 0 (0) & 6.45 (17.36) & 0 (0) & 0 (0) & 2.17 (4487.87) & 0 (0) & 0 (0) & 12.34 (87.88) \\
    53    & EKPC-MW \cite{dataset2:2018} & 1.76 (2138.5) & 0 (0) & 3.11 (2414.97) & 0 (0) & 3.07 (19.09) & 0 (0) & 0 (0) & 3.48 (5311.83) & 0 (0) & 0 (0) & 11.05 (127.79) \\
    54    & DEOK-MW \cite{dataset2:2018} & 2.75 (2180.97) & 0 (0) & 6.04 (2985.09) & 0 (0) & 4.33 (21.29) & 0 (0) & 0 (0) & 2.9 (5955.4) & 0 (0) & 0 (0) & 22 (181.54) \\
    55    & NI-MW \cite{dataset2:2018} & 1.14 (2324.72) & 5.47 (9722.64) & 0 (0) & 0 (0) & 8.31 (22.23) & 0 (0) & 0 (0) & 1.54 (5969.84) & 0 (0) & 0 (0) & 7.59 (171.02) \\
    56    & FE-MW \cite{dataset2:2018} & 0.56 (2174.84) & 0 (0) & 4.82 (3277.51) & 0 (0) & 3.98 (23.29) & 0 (0) & 0 (0) & 1.84 (6283.11) & 0 (0) & 0 (0) & 7.69 (205.07) \\
    57    & DOM-MW \cite{dataset2:2018} & 9.49 (3109) & 0 (0) & 7.92 (5726.12) & 0 (0) & 2.3 (34.28) & 0 (0) & 0 (0) & 19.39 (10538.77) & 0 (0) & 0 (0) & 39.57 (273.41) \\
    58    & DUQ-MW \cite{dataset2:2018} & 2.22 (3283.83) & 0 (0) & 2.17 (5653.23) & 0 (0) & 1.72 (33.12) & 0 (0) & 0 (0) & 2.9 (10571.89) & 0 (0) & 0 (0) & 18.83 (298.89) \\
    59    & AEP-MW \cite{dataset2:2018} & 0.74 (2194.24) & 0 (0) & 4.39 (6110.89) & 0 (0) & 1.26 (34.13) & 0 (0) & 0 (0) & 1.37 (10787.27) & 0 (0) & 0 (0) & 28.72 (212.06) \\
    60    & DAYTON \cite{dataset2:2018} & 1.89 (2685.16) & 0 (0) & 1.28 (6061.38) & 0 (0) & 0.81 (35.35) & 0 (0) & 0 (0) & 2.87 (10821.67) & 0 (0) & 0 (0) & 22.66 (141.72) \\
    61    & PJMW-MW \cite{dataset2:2018} & 10.35 (4356.42) & 0 (0) & 3.41 (7147.69) & 0 (0) & 1.14 (38.12) & 0 (0) & 0 (0) & 1.91 (12294.21) & 0 (0) & 0 (0) & 30.69 (287.08) \\
    62    & PJME-MW \cite{dataset2:2018} & 1.06 (3942.98) & 0 (0) & 6.54 (7209.55) & 0 (0) & 0.78 (39.52) & 0 (0) & 0 (0) & 4.19 (12292.27) & 0 (0) & 0 (0) & 28.32 (270.4) \\
 \bottomrule
 \end{tabular}}%
\label{uts12-models-smape}%
\end{table*}%

\setlength\extrarowheight{7pt}
\begin{table*}[htbp]
  \centering
   \caption{Symmetric mean absolute percentage error (smape) of all models along with execution time in seconds in parenthesis on \texttt{9} multivariate time series. The values displayed in each columns are "smape (total execution in seconds)". Smaller smape values indicate better performance. 0(0) value indicates a model did not finish in reasonable time w.r.t other models on the corresponding dataset and it was terminated. The datasets are ordered based on their sizes ranging from \texttt{106} data observations (walmart-sale) to \texttt{296,399} data observations (manufacturing). Models are ordered from left to right according to their average performance shown in Fig.~\ref{fig:mts_smape_rank}.}
\resizebox{\textwidth}{!}{%
    \begin{tabular}{rlrrrrrrrrrrr}
    \toprule
    Index & Datasets & AutoAI-TS & DeepAR & GLS   & RollingRegressor & WindowRegressor & Motif & PyAF  & Component & Prophet & NBeats & PMDArima \\
    \midrule
    1     & walmart-sale & 10.5 (73.01) & 0 (0) & 12.58 (154.52) & 0 (0) & 12.81 (495.47) & 12.2 (170.34) & 9.65 (27.98) & 13.85 (167.1) & 13.15 (3.14) & 12.54 (3449.82) & 10.69 (275.55) \\
    2     & nn5tn10dim & 22.09 (133.13) & 0 (0) & 26.39 (160.76) & 29.82 (782.85) & 27.96 (2385.45) & 27.74 (321.56) & 83.76 (79.2) & 34.32 (538.05) & 17.06 (4.26) & 43.15 (7002.02) & 36.4 (1062.53) \\
    3     & rossmann & 26.05 (229.6) & 25.44 (2598.25) & 25.11 (164.91) & 29.48 (836.61) & 35.1 (3036.38) & 26.43 (328.13) & 183.33 (81.25) & 45.17 (496.22) & 33.75 (4.5) & 38.57 (7313.24) & 37.77 (1583.5) \\
    4     & household & 29.88 (238.04) & 36.12 (2500.4) & 34.59 (145.36) & 33.75 (1962.04) & 32.64 (3734.37) & 36.2 (442.51) & 42.13 (73.09) & 32.66 (635.96) & 38.82 (5.77) & 49.54 (7695.69) & 36.07 (2117.13) \\
    5     & cloud & 18.65 (469.15) & 21.09 (1418.34) & 18.48 (79.09) & 18.51 (882.37) & 18.55 (3564.44) & 18.55 (802.14) & 19.98 (37.42) & 21.06 (378.68) & 66.16 (4.99) & 33.44 (10268.99) & 21.42 (3911.8) \\
    6     & exchange & 0.46 (2203.99) & 1.11 (4679.15) & 0.92 (209.17) & 0.73 (6196.52) & 0.98 (15298.96) & 1.04 (4320.55) & 0.46 (92.61) & 0.72 (2214.21) & 130.28 (162.22) & 1.96 (30572.98) & 0 (0) \\
    7     & traffic & 15.45 (8329.14) & 12.58 (11480.55) & 47.79 (385.54) & 29.39 (17566.16) & 16.1 (22338.26) & 39.36 (45978.43) & 200 (167.91) & 37.23 (26537.06) & 36.87 (47.72) & 21.77 (36359.13) & 0 (0) \\
    8     & electricity & 17.8 (23984.89) & 13.93 (16025.75) & 0 (0) & 0 (0) & 0 (0) & 0 (0) & 50.29 (190.53) & 0 (0) & 53.73 (485.89) & 14.35 (42015.16) & 0 (0) \\
    9     & manufacturing & 19.64 (80227.52) & 0.54 (74344.27) & 0 (0) & 0 (0) & 0 (0) & 0 (0) & 120.09 (399.71) & 0 (0) & 23.53 (2837.03) & 1.36 (115870.11) & 0 (0) \\
       \bottomrule
    \end{tabular}%
    }
  \label{tab:mts12-models-smape}%
\end{table*}%

\setlength\extrarowheight{8pt}
\begin{table*}[htbp]
  \centering
   \caption{Symmetric mean absolute percentage error (smape) of all \texttt{10} pipelines within the Autoai-TS along with execution time in seconds in parenthesis on on \texttt{9} multivariate time series. Smaller smape values indicate better performance. The datasets are ordered based on their sizes ranging from \texttt{106} data observations (walmart-sale) to \texttt{296,399} data observations (manufacturing). Pipelines are ordered from left to right according to their average performance shown in Fig.~\ref{fig:mts_inter_rank}.}
\resizebox{\textwidth}{!}{%
\begin{tabular}{rlrrrrrrrrrr}
    \toprule
    Index & Datasets & FlattenAutoEnsembler-log & WindowRandomForest & WindowSVR & MT2RForecaster & bats  & DifferenceFlattenAutoEnsembler-log & LocalizedFlattenAutoEnsembler & Arima & HW-Additive & HW-Multiplicative \\
    \midrule
    1     & walmart-sale & 9.88 (0.65) & 10.5 (4.01) & 10.08 (0.08) & 9.82 (0.25) & 10.76 (15.1) & 10.48 (0.05) & 11.27 (0.04) & 9.87 (0.16) & 9.47 (0.43) & 9.45 (0.4) \\
    2     & nn5tn10dim & 25.8 (5.48) & 22.09 (11.83) & 25.9 (1.55) & 35.94 (0.25) & 33.8 (28.4) & 33.19 (0.24) & 29.74 (1.14) & 36.21 (0.18) & 39.65 (0.71) & 39.66 (0.71) \\
    3     & rossmann & 32.72 (6.57) & 26.05 (13.96) & 37.43 (2.49) & 38.06 (0.26) & 33.09 (37.08) & 44.73 (1.41) & 27.37 (1.28) & 43.3 (0.19) & 39.12 (0.72) & 37.25 (0.84) \\
    4     & household & 35.93 (9.2) & 34.55 (22.37) & 29.88 (5.01) & 34.21 (0.27) & 35.79 (26.76) & 36.85 (0.59) & 36.12 (0.31) & 35.79 (0.2) & 38.4 (0.73) & 39.78 (0.78) \\
    5     & cloud & 18.84 (8.08) & 17.33 (96.85) & 23.88 (5.6) & 28.82 (0.09) & 26.54 (15.49) & 18.65 (0.61) & 39.7 (2.99) & 30.46 (0.13) & 28.32 (0.53) & 27.54 (0.56) \\
    6     & exchange & 0.95 (48.08) & 6.64 (110.28) & 9.87 (0.37) & 0.5 (2.73) & 0.45 (33.28) & 0.46 (5.21) & 0.8 (1.64) & 0.46 (0.95) & 10.8 (2.19) & 11.48 (2.46) \\
    7     & traffic & 15.45 (90.01) & 22 (350.82) & 87.84 (19.38) & 22.91 (2.67) & 58.97 (76.94) & 31.83 (30.99) & 22.6 (25.09) & 28.75 (4.83) & 25.31 (7.5) & 26.12 (8.34) \\
    8     & electricity & 17.8 (111.41) & 44.18 (387.24) & 22.61 (1710.94) & 27.33 (3.13) & 22.12 (124.54) & 20.09 (45.7) & 26.44 (36.19) & 29.06 (10.47) & 49.96 (14.03) & 88.72 (14.29) \\
    9     & manufacturing & 2.69 (23.78) & 19.64 (365.68) & 0.85 (15.15) & 2.44 (8.24) & 4.64 (383.35) & 1.44 (244.19) & 10.61 (34.47) & 1.76 (540.49) & 5.36 (565.38) & 56.36 (1756.66) \\
    \bottomrule
    \end{tabular}}%
  \label{tab:mts12-ppln-smape}%
\end{table*}%

\newpage
\bibliographystyle{ACM-Reference-Format}
\bibliography{paper} 

\end{document}